\documentclass[sn-apa]{sn-jnl}% APA Reference Style

\usepackage{graphicx}
\usepackage{multirow}
\usepackage{amsmath,amssymb,amsfonts}
\usepackage{amsthm}
\usepackage{mathrsfs}
\usepackage[title]{appendix}
\usepackage{xcolor}
\usepackage{textcomp}
\usepackage{manyfoot}
\usepackage{booktabs}
\usepackage{algorithm}
\usepackage{algorithmicx}
\usepackage{algpseudocode}
\usepackage{listings}

\usepackage{lmodern}

\begin{document}

\title[Article Title (short version)]{A Representationalist, Functionalist and Naturalistic Conception of Intelligence as a Foundation for AGI}

\author[1]{\fnm{Rolf} \sur{Pfister} \email{rolf.pfister@posteo.de}}

\affil[1]{\orgdiv{Munich Center for Mathematical Philosophy (MCMP)}, \orgname{Ludwig-Maximilians-Universit\"at M\"unchen}, \orgaddress{\street{Geschwister-Scholl-Platz 1}, \city{Munich}, \postcode{80539}, \country{Germany}}}

\abstract{
The article analyses foundational principles relevant to the creation of artificial general intelligence (AGI).
Intelligence is understood as the ability to create novel skills that allow to achieve goals under previously unknown conditions.
To this end, intelligence utilises reasoning methods such as deduction, induction and abduction as well as other methods such as abstraction and classification to develop a world model.
The methods are applied to indirect and incomplete representations of the world, which are obtained through perception, for example, and which do not depict the world but only correspond to it.
Due to these limitations and the uncertain and contingent nature of reasoning, the world model is constructivist.
Its value is functionally determined by its viability, i.e., its potential to achieve the desired goals.
In consequence, meaning is assigned to representations by attributing them a function that makes it possible to achieve a goal.
This representational and functional conception of intelligence enables a naturalistic interpretation that does not presuppose mental features, such as intentionality and consciousness, which are regarded as independent of intelligence.
Based on a phenomenological analysis, it is shown that AGI can gain a more fundamental access to the world than humans, although it is limited by the No Free Lunch theorems, which require assumptions to be made.
}


\maketitle

\section{Introduction} \label{sec1}

In recent years, extensive developments have taken place in the field of artificial intelligence (AI).
These include in particular generative AI approaches that use transformer or diffusion architectures and lead to contributions in many areas such as text and image generation \citep{788}, protein structure prediction \citep{789} and autonomous driving \citep{790}.
However, although these approaches achieve results that are considered impressive, they are unreliable and fail in many tasks that appear simple from a human perspective \citep{783, 785, 787}.
They also fail the more frequently the less similar the tasks are to those on which they were trained \citep{784, 786}.
Such weaknesses do not occur only in specific approaches, but constitute a general problem in the field of AI \citep{794, 795}.

As a consequence, AI applications can be used reliably in specific, controlled domains for which they have been designed and evaluated.
But AI applications often fail in more complex and practical tasks in which uncertainties occur; for instance, in autonomous driving \citep{791, 793}.
Currently, there is no artificial general intelligence (AGI), i.e., AI models that can solve a wide range of everyday tasks as reliably as humans can \citep{796}.
The development of AGI is considered a desirable goal, as AGI could relieve humans of tasks they do not want to perform.
Furthermore, with AGI, a single AI model could be used for all types of tasks instead of having to develop a separate model for each specific use case, as at present.

The aim of the article is to identify and analyse principles that have to be considered for the creation of AGI.
The analysis focuses in particular on understanding intelligence and how AGI can perceive and interpret a world in such a way that it can reliably fulfil a wide range of goals.
The analysis is not about the evaluation of a specific AI approach such as symbolic, embodied or generative AI, but about the foundational characteristics of AGI.

Section \ref{sec2} analyses different conceptions of intelligence and concludes that intelligence is the ability to create novel skills that allow one to achieve goals under previously unknown conditions.
Section \ref{sec3} discusses the role of prediction and the necessity for intelligence to be based on assumptions about the world in which it is to be applied.
Section \ref{sec4} is concerned with perception, its indirect and representational nature, and its distinction from conscious experience.
Section \ref{sec5} examines the nature of representations and shows that they are an inherent aspect of grasping a world to determine goal-directed actions.
Section \ref{sec6} explores how a world is grasped and, based on the phenomenological approaches of Heidegger and others, outlines the dichotomy between a world itself and the interpreted conception of it.
Section \ref{sec7} analyses the conceptions of meaning and understanding and argues for a functional definition of them, which allows for a naturalistic interpretation of intelligence that does not require assumptions of mental features such as consciousness.
Section \ref{sec8} describes how intelligence utilises reasoning methods such as deduction, induction and abduction, as well as abstraction and classification for the development of world models.
Section \ref{sec9} discusses the assessment of world models on the basis of their functional usefulness, i.e., viability, rather than their depiction of truth, and discusses their constructivist character, which results from the uncertainty and contingency of the reasoning methods.
Section \ref{sec10} addresses the subjective perspective through which an agent perceives a world and examines the numerous interrelations between an agent and the rest of the world.
Section \ref{sec11} concludes with an overview of the approach developed in the article, outlining the foundational characteristics that have to be considered to enable the creation of AGI.

\section{Skills \& Intelligence} \label{sec2}

For the development of AGI, it is important to understand its nature precisely.
This includes in particular the concept of intelligence.
Human intelligence is explained by the Cattell-Horn-Carroll theory as an interaction between crystallised intelligence and fluid intelligence \citep[pp. 73-75]{797}:
Crystallised intelligence consists of several broad cognitive abilities, such as reasoning, processing visual information, and remembering information.
Fluid intelligence is a general ability whose performance affects all broad abilities and describes the general cognitive capacity.
In the field of AI, a variety of definitions of intelligence are used \citep{798}, which can be broadly categorised into two groups:
Process-oriented definitions name required abilities such as learning, abstraction, logical thinking, and problem solving.
Result-oriented definitions focus on the outcome and define intelligence as the ability to achieve specific goals; for instance, to adjust to an environment, to create products, or to grasp truths.

To determine whether an AI approach is intelligent, it is usually tested on tasks that fulfil the requirements of the definitions.
In the course of the history of AI, numerous tasks whose solutions were assumed to require extensive cognitive abilities, and therefore intelligence, were proposed.
The proposed tasks included for example playing chess, playing Go, image recognition, translating texts, or creating meaningful texts.
However, when AI approaches were able to solve any of the problems, they were considered not intelligent.
One reason for this is that the methods used by the approaches to solve a task, for example trying out a large number of possibilities, are not considered intelligent.
It is also argued that the tasks are not solved by the intelligence of the AI approaches but by the intelligence of the programmers embedded in the approaches.
Moreover, it is argued that an approach cannot be intelligent if it can solve a task but fails if the task is modified; a problem that concerns many approaches.
This leads some to conclude that AI approaches are making major progress in terms of performance but not in terms of intelligence (\citealp[pp. 396-404, 421-423, 434]{799}; \citealp[pp. 7-9, 16f]{674}).

\citet[pp. 3-7]{674} explains this contradictory development by the fact that two different interpretations of intelligence are used and that they are not distinguished sufficiently clearly.
The first interpretation understands intelligence as a collection of task-specific skills, as advocated by Darwin and Minsky, for example.
The second interpretation understands intelligence as the ability to create novel skills for solving tasks, as advocated by Turing and McCarthy, among others.
Accordingly, while the first interpretation classifies solving tasks known to an AI approach as intelligent, the second interpretation classifies solving tasks hitherto unknown to an approach as intelligent.
\citet[pp. 18-20]{674} argues that the first interpretation of intelligence as task-specific skills is misleading because it does not describe intelligence but only its output:
Skills are specific solutions to specific problems that are created by intelligence but that are not intelligence itself.
In contrast, the second interpretation describes intelligence as a process, as an ability that creates skills.

A further reason in favour of the second interpretation of intelligence is that only that one is suitable for the development of AGI.
This, as skills can be applied to specific tasks for which they were created, i.e., tasks that are known and well-defined, such as mastering games.
But skills cannot be reliably applied to tasks outside the well-defined domain for which they were created:
Skills do not include specifications on how to handle unfamiliar conditions\footnote{
In the field of AI, conditions are often called states.
The two terms are used interchangeably in this article.
} that occur outside the well-defined domain.
Everyday tasks from the human domain, which AGI is supposed to solve, often have unfamiliar conditions:
The future development of the world is only partially predictable for humans -- and thus also for skill-based AI approaches created by humans -- and future conditions remain partially unknown.
Accordingly, AGI cannot be realised via a skills-based approach, as it would not be able to handle the constantly arising new, unknown conditions.
Instead, AGI must be able to create novel skills to cover the unknown conditions, i.e., AGI must be able to fulfil the second interpretation of intelligence.

The foregoing considerations allow for a more precise definition of skill and intelligence:
A skill is the ability to achieve a specific goal under specific known conditions.
Intelligence is the ability to create novel skills that allow to achieve goals under previously unknown conditions.
As such, intelligence is also a skill:
it is a skill that allows to create other skills.
Intelligence is not a fixed ability that is only either present or absent, but one that can also be stronger or weaker:
An agent is the more intelligent, the more efficiently it can achieve the more diverse goals in the more diverse worlds\footnote{
In this definition, a world is seen as a self-contained and independent system that can have different conditions, some of which are accessible to the agent and some of which may be manipulable by the agent.
Instances of individual worlds are the universe in which humanity is situated, games such as Go and computer games, and mathematical and logical systems.
}
with the less knowledge.
Knowledge is understood pragmatically here:
It does not have to be true statements about the worlds, but it includes all the information the agent has, including skills.
The negative consideration of knowledge in the definition of intelligence entails that only the ability to generate skills but not skills themselves falls under intelligence.
The definition thus corresponds to the second interpretation of intelligence discussed by Chollet above and excludes the first interpretation.
Simply put, intelligence describes how well an agent can achieve goals in novel, unknown conditions.

The juxtaposition of the application of existing skills on the one hand, and the generation of skills, i.e., intelligence, on the other, reveals a fundamental relationship between the two:
Tasks can be solved either by skills or by intelligence.
This means skills and intelligence can be substituted for each other, provided that all conditions are known.
Intelligence is only necessary to the extent conditions are unknown or skills are not available for other reasons; for example, because skills cannot be provided for all possible known conditions.
Beyond that, the assessment of the degree of intelligence is abstract in that it does not permit a quantitative assessment without further specification of how this is to be carried out.
For example, the assessment does not describe how exactly efficiency or diversity are quantified, or how the individual factors are weighed against each other.
However, the provision of such specifications is not necessary for the further course of the article.
\citet[pp. 27-42]{674}, who provides a measurable definition of intelligence, states that many possible ways of measuring intelligence may be valid.
Which specific quantitative valuation is the best requires further research and may depend on epistemic as well as ontological assumptions.

The above definition describes intelligence as an ability of an agent.
An agent is defined in this article as a system that is able to perform specific actions depending on specific conditions to achieve specific goals.
Understanding AI approaches as agents is a fundamental perspective within the field of AI \citep[pp. 7, 21f]{1007}.
With regard to AGI, the aim is to develop an AI agent that is intelligent, i.e., an agent that is able to fulfil goals under partially unknown conditions.
The goals are specified by the creator of the agent, i.e., by humans.
They can be of a more specific nature, such as controlling a vehicle, or of a more general nature, such as developing scientific theories.
For an agent, skills provide specifications under which conditions which actions are appropriate to achieve a specific goal; for example, in which chess position which move is appropriate to win the game.
As shown above, if an agent encounters conditions that are at least partially not covered by skills, the specifications provided may be insufficient to achieve its goals; this, because it is uncertain how the uncovered conditions will affect the achievement.
Consequently, the agent must utilise intelligence to create a skill, i.e., provide the specifications on how to achieve its goals under the unknown conditions.
To determine which possible actions are appropriate, the agent must determine how they affect the achievement of the goal.
This means that the agent has to make a prediction:
It has to determine how a specific action influences the achievement of its goals without performing the action.

\section{Prediction \& Assumption} \label{sec3}

A prediction is a specification of unknown conditions.
Conditions can be unknown to an agent, for example, because they occur in the future or because the agent cannot perceive them for other reasons.
To be successful, a prediction requires knowledge of the world, i.e., of some of its conditions.
Furthermore, a prediction requires knowledge of how the conditions of the world develop; i.e. it requires a model of the world that describes the development of the conditions to be predicted on the basis of the current conditions of the world.
The applicability of such a world model requires that the world is subject to at least some regularities.
If all conditions of a world were irregular, for example because they were completely random, there would be no regularities that could be part of the world model and used to specify unknown conditions.
Consequently, predictions -- and therefore intelligence -- can succeed only in worlds that exhibit at least some regularities \citep[cf.][pp. 1300f]{766}.

The No Free Lunch (NFL) theorems show that across all possible optimisation problems any algorithm has the same average performance as every other.
Consequently, there is no algorithm that is better than others at solving all optimisation problems:
If an algorithm performs better than another on one set of optimisation problems, it performs worse than the other on the set of all other optimisation problems (\citealp[pp. 69-71]{801}; \citealp[pp. 4f]{802}).
This can be seen as a counterargument to the formalisation of intelligence:
Intelligence is about solving unknown optimisation problems with above-average performance, but the NFL theorems indicate that there cannot be such an algorithm.
However, as shown above, intelligence can only be beneficial in worlds that have at least some regularities.
This means that intelligence does not have to be adapted for all possible optimisation problems but only for the subset of optimisation problems that occur in worlds with regularities \citep[cf.][pp. 402f]{799}.
Consequently, it is possible to find an algorithm that performs better than others on this subset of problems -- and worse on the remaining optimisation problems of completely irregular worlds.

For an algorithm to be better than others on a subset of optimisation problems, the characteristics of the subset must be incorporated into the algorithm \citep[pp. 71f]{801}.
In the case of intelligence, the algorithm has to be optimised with respect to regularities \citep[cf.][pp. 1300f]{766}.
The regularities considered are thereby not necessarily truths of the worlds but assumptions.
The formalisation of intelligence thus faces a dilemma in regard to determining to what extent regularities -- and possible other assumptions -- should be considered:
The more assumptions are considered, the smaller the subset of optimisation problems covered and the more performant the algorithm, all else being equal.
However, the more assumptions are considered, the greater the chance that they do not correspond to the worlds to which the algorithm is applied, and its performance decreases accordingly.

\section{Perception \& Experience} \label{sec4}

Skills and intelligence both require knowledge of at least some conditions of a world to determine appropriate actions to achieve a goal.
Conditions can be determined through perception.
For example, humans and animals can perceive stimuli that can be divided into three different types:
Chemical stimuli include molecules and are experienced as odour and taste; mechanical stimuli include forces transmitted by matter and are experienced as touch, sound and heat; electromagnetic stimuli include electrical and magnetic radiation and are experienced as vision, for example.
Stimuli are detected by receptors located in sensory organs, such as eyes.
Together with the nerves that transmit and process their signals, sensory organs are referred to as sensory systems.
For example, the human visual system includes the eyes, the connected nerves, and the visual cortex of the brain \citep[pp. 7-11, 191, 213f]{1006}.
Sensory organs can vary in performance, e.g., regarding the type and detail of stimuli that can be perceived.
Eyes, for example, can be divided into four stages of functional efficiency:
In the first stage only the presence of light can be perceived, in the second stage also the rough direction from which the light comes.
The third stage allows the perception of more detailed directions and therefore contrasts; and the fourth stage, through the use of lenses, allows sharp spatial vision at distance \citep[pp. 2837-2843]{804}.

Sensory organs and their performance thus represent a limitation as to which conditions of a world can be perceived and in what detail.
The limitations of sensory organs can lead not only to a lack of perception but also to distorted perceptions.
For example, flickering light is experienced as continuous light above a particular flickering speed due to the limited temporal resolution of the sensory system.
Another example is the human perception of the sky as blue:
This occurs because the shorter the wavelength of light is, the more it is scattered and therefore the better it is perceived.
Accordingly, the shorter-wave, blue light component of sunlight is scattered more strongly in the Earth's atmosphere than the longer-wave, red light component.
However, the violet light component is even shorter in wavelength and is therefore scattered even more strongly.
Yet, as human receptors perceive blue light more strongly than violet light, the sky still appears blue from a human perspective \citep[p. 253]{800}.
Moreover, many optical illusions demonstrate that conscious experiences do not correspond to what is perceived and some optical illusions persist even when one is aware of their incorrectness \citep[e.g.][pp. 40-50, 127-134]{989}.

In subjective human conscious experience\footnote{
These experiences are often referred to as qualia.
For a detailed discussion of qualia, see \citet{845}.
},
stimuli seem to be experienced directly, as if one experiences the stimuli themselves.
Nevertheless, the relationship between stimuli and human experiences can be indirect and varying.
Some sensory experiences are not generated directly by specific stimuli, but are generated by sensory systems.
The colour yellow, for example, is perceived as a direct and genuine experience of a stimulus, just like the colour red.
However, the colour yellow is not experienced because a colour receptor for yellow light is activated.
Instead, it is experienced when green and red colour receptors are activated simultaneously \citep[cf.][pp. 88-91]{805}.
Hence, although the colour yellow appears as a direct perception of a stimulus, it is a generated experience without a corresponding stimulus of its own.

Furthermore, an individual stimulus can be experienced as perceptions of several sensory systems simultaneously.
Synaesthetes experience, for example, sonic waves not only as sound but also visually as colours, whereby the experienced colours can differ depending on the person.
Equally, their perception of light can lead not only to visual experiences, but also to experiences of taste \citep[pp. 50-56]{806}.
Conversely, stimuli of different types can trigger the same sensory system.
For instance, capsaicin in chilli and menthol in mint produce the experience of heat and cold respectively, as the molecules activate temperature receptors \citep[pp. 746f, 751]{807}.
The joint processing of stimuli of different types within a sensory system is widespread among animals:
Platypuses combine signals from receptors for electric fields and mechanical forces, mosquitoes have neurons that react simultaneously to both temperature and chemicals, and migratory birds process the perception of both light and magnetic fields in the visual centre \citep[pp. 314f, 323f]{1006}.
In addition, different sensory systems can be activated by the same stimuli.
For example, odour and taste are partially activated by the same chemicals, such as esters and amino acids.
Odour and taste thus do not differ primarily in that they perceive different types of stimuli; rather, their difference is functional:
Reactions to taste are reflexive and innate, whereas those to odour are learnt and depend on experience \citep{808}.

In summary, the same stimuli can trigger different experiences, and, conversely, different stimuli can trigger the same experiences.
This shows that human subjective conscious experience is not a direct and unaltered experience of stimuli but an indirect and varying one.
One of the reasons for this lies in how stimuli are perceived.
In their basic functioning, all sensory systems are structured in the same way, regardless of the type of stimuli they perceive:
A stimulus triggers in complementary receptors a chemical or electrical reaction that leads to an electro-chemical activity of the receptors' neurons, which in turn results in neural activity in the sensory system \citep[pt. 2]{1008}.

Odours, for example, are experienced when receptors are activated by specific chemical molecules.
When molecules activate corresponding receptors, the receptors send a signal and release or destroy the molecule.
However, there is not a specific receptor for every particular odour.
Instead, many types of molecules activate several different receptors at once, and depending on which receptors are activated simultaneously, different odours are experienced.
The characteristics of the receptors and their interaction depends on genes; for instance, the OR7D4 gene determines whether androsterone, a male sex hormone, is experienced as repulsive, vanilla-scented, or odourless \citep{809}.
Visual perception relies on the same process, except that the relevant receptors, opsins, do not hold and repel molecules but are permanently connected to a chromophore molecule.
When a photon hits a chromophore molecule, its energy changes the shape of the molecule, which in turn leads to neural activity of the receptor \citep[pp. 3f, 11f]{810}.
In hearing, hair cells are involved which, depending on the movements caused by sonic waves, release chemical substances that then lead to neuronal activity \citep[ch. 9]{1008}.

Common to all these and other sensory systems is that stimuli themselves are not retained \citep[pp. 115f]{977}.\footnote{
The electrical sense, which allows to perceive electrical fields, could be considered an exception: Both stimuli and neural activity are electrical in nature.
Yet, here too, the electrical stimulus is not continuously preserved, but its presence triggers chemical activity, which in turn leads to neural electrical activity that differs from that of the input \citep{811}.
}
Instead, a stimulus leads to a neural activity of an electro-chemical nature, which is dependent on various aspects of the stimulus; in the simplest case on its presence.
The conscious experience of perception in humans is therefore not a direct experience of stimuli themselves, but is based on neural electrochemical activities caused by the stimuli.
Overall, this shows that perception in humans and animals is the ability to convert stimuli, i.e., conditions of a world into neural activity.
Generalised, perception can be defined as the ability to form states in dependence on conditions of a world.
As such, the formed states are representations of conditions of the world.
However, as shown above, the connection between the conditions of the world and their representations can be incomplete, ambiguous, and inaccurate due to the limitations of the sensory organs.

Since sensory organs provide only representations and not the stimuli themselves, representations can also be provided by other means.
Accordingly, although intelligence requires knowledge of at least some conditions of a world, this does not necessarily have to be obtained through perception.
Instead, knowledge can also be provided in other forms, such as a database.
Examples of worlds for which knowledge is provided in this way, both for AI systems and for humans, are games or logical and mathematical systems for which axioms are provided instead of perceptions.
Nevertheless, in principle, a comprehensive and precise perception is favourable:
The more conditions of a world are known, the more precisely it is possible to determine which actions are appropriate, all else being equal.
Furthermore, perception allows one to continuously obtain conditions of a world, allowing, for instance, the consequences of actions or previously unknown states to be determined.

\section{Representation} \label{sec5}

\citet[pp. 149-158]{734} describes a robot called Herbert, which is an intermediate result of his approach to creating intelligent systems.
The robot can move around in a regular office environment to collect empty soda cans.
It is controlled by fourteen activity modules, each designed for a specific function; for example to avoid obstacles, to recognise tables, or to grasp objects.
Accordingly, the robot is not based on classic AI approaches, such as symbolic AI reasoning systems, or neural networks.
Instead, the robot is controlled by the activity modules and their interaction.
The modules are interconnected and different modules take over control at different times depending on their states.
For example, by default the robot wanders around.
Yet, if the avoidance module recognises an obstacle, it takes over and changes direction.
Equally, when a soda can is discovered, the grasping module takes over to stop the robot and to grasp the can.\footnote{
Each activity module is based on a hardwired fixed-topology-network of simple finite state machines.
As such, each module represents a specific skill.
The robot is not able to adapt to novel, unknown circumstances; for example, it would not be able to learn to grasp soda cans of different shapes or bottles.
Consequently, the robot fulfils only the first interpretation of intelligence outlined in Section \ref{sec2}, but not the definition of intelligence advocated in this article.
}

\citeauthor{734} (\citeyear{734}, pp. 148f, 140, 154; \citealp[cf.][pp. 175-180]{998}) takes the seemingly strict position that the approach does not rely on representations because there are no "tokens which have any semantics that can be attached to them".
Subsequently, \citet[pp. 18-20]{751} takes a more nuanced position, which does not entirely deny the presence of representations, but rejects the presence of "explicit representations", "symbolic representations", and "traditional Artificial Intelligence representations schemes".
The divergence seems to be primarily due to an insufficient differentiation between various kinds of representations, and the attempt to demarcate from traditional AI approaches that are based on logical systems involving natural language.
At least implicitly, \citet[p. 157]{734} functional description of the robot refers to representations: "For instance the grasp behaviour can cause the manipulator to grasp any object of the appropriate size seen by the hand sensors."
In other words, in case the hand sensors perceive stimuli typical for a soda can, the grasp behaviour module sends a signal to the manipulator.
This signal is thus a representation of the perception of a soda can; as Brooks states it: "aspects of the world are extracted".
The same applies to other modules:
For example, the ultrasonic sensors of the obstacle module send a signal when they are activated by an object.
The signal thus represents a state of the world in which there is something in front of the ultrasonic sensor that activates it.

Representations vary in how vaguely or specifically they describe states of a world and how simple or complex they are.
An example of rather simple, yet functional representations provides the water flea Daphnia.
Its visual sensory system is not able to perceive details but can only recognise the presence of four different wavelengths of light.
Depending on the wavelength of the perceived light, one of four different types of opsins is activated.
Each type of opsin sends a specific signal, which thus represents the presence of light of the corresponding wavelength.
Based on these representations, specific actions are triggered.
For example, the signal representing the presence of green light, which indicates the presence of food, triggers the action to swim in the direction of the light.
Similarly, the signal representing the presence of UV light, which indicates damaging UV radiation, causes the insect to swim away from the light source.\footnote{
In detail, the reactions are more complex than described here. For example, the insect's reactions are also influenced by the circadian rhythm and genetic dispositions.
In addition, very intense green light also causes the insect to move away from the light source rather than towards it.
For reasons of illustration, these additional influencing factors are not taken into account here.
}

In both examples, the signals represent very simple states, namely the presence of light of particular wavelengths.
However, representations can also be more vague as well as more complex.
For example, the representation of a forest has a higher level of complexity, as it includes a larger number of trees, other plants, and animals as well as a terrain.
In addition, it has greater vagueness, as forests can include a wide variety of plant and animal species and can be of different kinds, all aspects that are not specified in the representation \citep[cf.][pp. 81f]{998}.
Consequently, there are many different possible sets of states of the world that can lead to the same representation, and for all of which the representation stands accordingly.
The complexity of a representation is of a gradual nature and depends, among other things, on the number of possible states represented, as well as on the variety in which they can be combined.

How easily and precisely a representation can be described depends not only on its complexity but also on the availability of suitable linguistic expressions.
For example, the German term ‘Regenschirm’ refers to an umbrella that is used specifically to protect against rain; consequently, the representation of such an umbrella is easier to describe in German than in English.
While the complexity of a representation is an inherent property, its describability depends on the language used and is consequently independent of the representation itself.
Accordingly, \citet[pp. 148f]{734} distinction between ‘implicit’ representations and ‘explicit’ or ‘symbolic’ representations cannot be upheld:
Explicitations and symbolic connotations of representations are only assignments, but not inherent aspects of the representations.

Based on the above, representations can be defined in the following way:
A representation is a state that is dependent on one or more other states.
Representations do not have to reflect other states completely, but can reflect only specific aspects of them.
For example, the representations of the water flea Daphnia indicate only the presence of light of a particular wavelength, but not the polarisation or spatial distribution of the light.
Moreover, representations can be indeterminate insofar as they can stand for several possible combinations of states, as the example of forests shows.
\citet[p. 182]{998} argues that a definition of representations based only on dependencies is too broad; instead, representations "must be used as stand-ins by someone or something to count as representations".
However, like the assignment of linguistic terms to representations, their use is something extrinsic -- whether a representation is used or not is not part of the representation itself.
As an illustration can serve a water flea whose opsins function normally but whose nervous system fails to process signals and thus to trigger actions.
In that case, the representation of the perceived light generated by the opsins is not used; however, it is the same representation that a functional water flea would have that would use the representation.

\section{Phenomena \& Appearances} \label{sec6}

Based on the considerations in the last section, it therefore appears that representations are a fundamental component in the implementation of intelligence, as they provide information about states of a world.
This view is widely held, particularly in the field of AI, where representations are assumed to be necessary for human and animal behaviour, as well as for AI approaches \citep[pp. 31, 76-78, 226f]{1007}.
However, \citet[pp. 249-251]{701} argues on the basis of the relevance problem that AI approaches which are applied in dynamically changing worlds cannot be based on representations:
AI approaches have to determine in specific situations which states of a world are relevant and which consequences result from changing states.
Yet, representations of states are meaningless and, as part of this, provide no information about their significance.
The meaning of a state of a world could be determined by knowing the concrete situation in which it occurs.
For example, the significance of a red traffic light for cars depends on whether one is participating in the situation as a driver or as a pedestrian, as well as on the direction one intends to take.
In order to determine the meaning of a represented state, an AI approach would therefore have to determine the situation in which it is applied.
But to do so, it would have to determine which states of the world form the situation, in other words, which states are relevant.
This leads to an infinite regress that cannot be overcome, as both the meaning and the situation can be determined only on the basis of the other.
Although Dreyfus' criticism is directed against symbolic AI approaches, he also applies the argument to other approaches that use explicit rules to manipulate representations \citep[p. 99]{1011}.

To overcome the relevance problem, \citet[pp. 252-255]{701} argues, AI approaches must not be based on representations but must be able to perceive solicitations:
"In coping in a particular context, say a classroom, we learn to ignore most of what is in the room, but, if it gets too warm, the windows solicit us to open them. We ignore the chalk dust in the corners and the chalk marks on the desks but we attend to the chalk marks on the blackboard. We take for granted that what we write on the board doesn't affect the windows, even if we write, 'open windows,' and what we do with the windows doesn't affect what's on the board" \citep[p. 263]{701}.
In conclusion, solicitations arise from concrete situations and provide meaning.
They disclose the world and offer a flexible response based on the significance of the current situation.
In contrast to representations, which are part of an AI approach and are only assigned to a world, solicitations are the world itself \citep[p. 249]{701}.
Accordingly, the meaningful is provided to an agent by the world, and appropriate actions do not have to be determined by the agent, but are offered as dispositions to respond to the solicitations of situations \citep[p. 367]{776}.\footnote{
This perspective is similar to the non-representational and non-computational account of \citet[pp. 119-121]{1012}:
Gibson, speaking of affordances instead of solicitations, argues that possibilities for action are offered to animals by their environment.
Affordances are not part of the agent but part of the environment, and perception is not about perceiving and processing information but about receiving guidance for action.
}

Whereas Dreyfus' account is intended to overcome the relevance problem, it requires a strong ontological commitment:
The approach presupposes the existence of solicitations for each agent in each situation.
It is not clear where the solicitations originate and what nature they are.

Furthermore, Dreyfus' account is in contradiction to the above findings from the analysis of Brooks' robot Herbert.
\citet[pp. 249f]{701} does not see Brooks' approach as a solution to the relevance problem, since the approach is not able to learn and thus cannot deal with changing meanings in novel situations.
Nevertheless, he considers Brooks' approach to be an important advance, as it is not based on representational, symbolic AI approaches, but on activity modules that react directly to the environment.\footnote{
A similar assessment is provided by \citet[pp. 175-180]{998}.
}
However, as shown above, Brook's approach is based on representations that, although not annotated with symbols or expressions of natural language, are processed according to explicit rules.
This raises the question of whether Dreyfus' approach of solicitations is, at least partially, based on representations, too.
\citet[pp. 333-342]{746}, who also regards Brooks' approach as a major advance, presents an account which comes close to the one of Dreyfus, but relies on representations.
\citet[p. 263]{701}, however, explicitly rejects this account, arguing that any representational state precludes meaning; instead, it is necessary to directly sense and respond to the world.
This raises the question, if solicitations are entirely non-representational, how can they be recognised by AI approaches, as well as by humans and animals, if not by means of their representational sensory systems (\citealp[cf.][pp. 249-251, 256-265]{701}; \citealp[pp. 364-369]{1013}).

Beyond that, Dreyfus' approach seems not suitable as a basis for intelligence.
According to \citet[p. 250]{701}, with increasing experience, we are presented with more and more finely discriminated situations that solicit increasingly detailed responses.
As background know-how is refined, states of the world take on more and more significance.
Additionally, \citet[p. 263]{701} explains:
"[W]henever there is a change in the current context we respond to it only if in the past it has turned out to be significant, and when we sense a significant change we treat everything else as unchanged except what our familiarity with the world suggests might also have changed and so needs to be checked out."
Yet, with this statement, Dreyfus does not describe how the world provides solicitations, and thus meaning in situations.
Instead, he describes how we cope with situations by applying our existing knowledge -- in other words, by applying skills.
In contrast, Dreyfus' approach does not allow for the application of intelligence, as it does not explain how we are able to perceive the meaning and significance of states of the world that are unknown to us.\footnote{
\citet[p. 264]{701} describes that we are made aware of new situations and states by having our attention drawn, "summoned", to them.
However, this can only explain how to switch from one skill to another, but not how to create new skills that can address new, unknown situations and determine the meaning of unknown states.
}

Overall, therefore, it seems that Dreyfus' account is not suitable as a foundation for intelligence.
Nevertheless, the question arises as to whether insights can be gained from his approach and the underlying considerations that are helpful for understanding intelligence and for the creation of AGI.
Dreyfus draws largely on considerations from phenomenology, in particular from the works of Heidegger and Merleau-Ponty.

Phenomenology focuses on phenomena and appearances and their conditions of possibility.
Appearances, i.e., conscious experiences of phenomena, play an important role in phenomenology, since they are the most immediate to which one has access \citep[cf.][pp. 45f]{1029}.
Yet, phenomena are not, as is often mistakenly assumed, equal to the immediate appearances that one consciously experiences \citep[pp. 11, 21-23, 251]{983}.
Instead, phenomena are the essential structures that characterise appearances.
Phenomenology is therefore not primarily concerned with the investigation of appearances as such, but with the investigation of phenomena, of appearances as their correlates, and of the connection between the two \citep[pp. 23-28]{983}.
As \citet[pp. 36f]{1001} describes: just because phenomena are proximally and for the most part not given, there is a need for phenomenology.
He argues, the idea of grasping and explicating phenomena in a way which is original and intuitive is directly opposed to the naïveté of a haphazard, immediate, and unreflective beholding.
The aim of phenomenology is thus not the description of subjective content of experience, but the determination of necessary and invariant features and the answering of questions related to truth, reason, reality, being, ontology, science, and objectivity \citep[p. 28]{983}.

However, this does not mean that appearances and phenomena are distinct from each other.
Phenomena are not represented by appearances but unfold in them; appearances are thus part of phenomena \citep[pp. 23-28]{983}.
Consequently, no distinction can be made between subjective experience on the one side and objective reality on the other.
Phenomenology is thus directed against the assumption of scientific realism that there is an objective reality that can be understood by removing all subjective elements of perception.
Instead, the objective, necessary, and invariant features can be understood only if conscious experiences, i.e., appearances, are part of the investigation \citep[pp. 108-114]{983}.
Phenomenology hence reflects that science is carried out by someone and thus from a specific theoretical stance, which has its own presuppositions and origins.
These presuppositions and origins need to be examined, which is why phenomenology is concerned, for example, with what the primitive modes of understanding are that precede beliefs in objectivity and how objectivity is constituted.
In this way, phenomenology aims to provide a new epistemological foundation for science \citep[pp. 23-28]{983}.

One of the main representatives of phenomenology is Heidegger, to whom Dreyfus refers most strongly.
\citet[pp. 10f]{1014} emphasises that our primary relationship to being, i.e., to the world in its entirety, is not in theoretical contemplation and investigation, but in immediate experience:
For example, we do not hear a sequence of sounds, but we experience the closing of a door.
In order to hear the sounds, we first have to reflect ourselves out of the situation and listen abstractly.
The sounds thus not only represent an abstracted and hence reduced view, their characterisation is also based on theoretical assumptions, such as the existence of an objective world.
Such views are therefore not suitable as a foundation for scientific investigations and insights, as they are already incomplete and may be based on erroneous assumptions.
Instead, investigations have to start in the immediate experience:
Only in the realisation of existence, called being-in-the-world, phenomena have the opportunity to reveal themselves and to disclose meaning.
For example, we only recognise the meaning of music when we not only perceive it as a sequence of sounds but experience it as music (\citealp[pp. 171-179]{1015}; \citealp[cf.][pp. 177-182]{983}).

Heidegger's change from the assumption of an objective world to immediate experience entails a different understanding of the role of cognitive abilities.
These no longer serve to establish the relation between the self and the world.
Instead, the world unfolds within the being-of-the-world, and relations between phenomena result from this.
Cognition thus becomes a secondary modification of being-in-the-world and is only possible because that is already present \citep[p. 178]{983}.
Heidegger's approach is thus closely related to Dreyfus' approach.
In consequence, Heidegger's approach faces the same limitations as Dreyfus' in relation to skills:
The perception of appearances is immediate, but at the same time it is already a matter of specific interpretation; for example of sounds as music \citep[p. 8]{983}.
An assessment that is also supported by \citet[pp. 801f]{1016}, who states:
It belongs to experience that something appears in it, but the interpretation makes up what we call appearance -- be it correct or not, anticipatory or exaggerated.
Heidegger's approach is therefore in some respects more direct and less presuppositional than, for example, scientific realism, but at the same time it is also based on interpretations and thus on assumptions.

In Heidegger's works, a clear change in perspective can be recognised between his earlier and later writings, which he himself describes as a turn.
While all the above considerations derive from his earlier writings, \citet[pp. 173-185]{1020} argues in his later writings that an understanding of the world requires, moreover, an engagement with the openness of unconcealment.
In his earlier writings, to which Dreyfus refers, Heidegger is concerned primarily with the question of how one can experience the world directly by being-in-the-world.
In his later writings, Heidegger focuses more on being, which, he states, he did not sufficiently consider in his earlier writings, as he focused too much on being-in-the-world \citep[pp. 49f]{1017}.

Being can be contrasted with being-in-the-world:
In being-in-the-world, in the realisation of existence, phenomena manifest themselves in the form of appearances and have meaning.
In the experience of being, however, one transcends concrete existence and experiences phenomena without interpretation:
One experiences the inexhaustibility of the world and discovers the possibility that existence can also be different.
In concrete terms, being contains all practised, all conceivable, and all as yet inconceivable possibilities of being-in-the-world.
At the same time, being eludes definition; the moment it is defined, it becomes being-in-the-world and is no longer being.
Accordingly, the experience of being is not present when one is trapped in one's own being-in-the-world; i.e. when one experiences the world in a specific way that is determined by a particular interpretation and from which one cannot free oneself.
Being can thus be understood as a game of possibilities that allows one to see the world as it is without interpretation, without a particular world view \citep[pp. 153-169]{1018}.
Yet, this world is not graspable since it consists of that very play of possibilities without ever adopting a particular one \citep[pp. 224-227]{1019}.
Metaphorically speaking, concrete realisations of being-in-the-world, such as religions and cultures, can be understood as fragile rafts that humans build on the open sea and on which they drift through time for a while, while modifying and sometimes rebuilding the rafts \citep[pp. 341-343, 406-409, 473f]{999}.\footnote{
A similar position is held by \citet{831}.
For a comparison of both positions, see e.g. \citet[pp. 66-68, 276-278, 336-343]{999}.
}

Heidegger's earlier approach and his later approach show clear parallels to the concepts of skills and intelligence.
Heidegger's earlier approach, to which Dreyfus also refers, describes the perception of and interaction with a world from a specific perspective, i.e., skill, whereby things have a specific meaning.
One example of this is Brook's robot Herbert, whose task is to identify objects shaped like soda cans and to pick them up.
Another example is the game of chess, in which pieces have a specific function and the game follows specific rules.

Heidegger's later approach, by contrast, describes the experience of phenomena without them being subject to any particular interpretation.
This corresponds to situations in which one is confronted with unknown states of the world and in which one therefore has to apply intelligence to be able to interpret them.
Both Heidegger's later approach and intelligence are therefore concerned with assigning meaning to uninterpreted phenomena in order to gain new insights.
Thereby it becomes apparent that Heidegger's uninterpreted experience of phenomena is subject to the same restrictions as intelligence with regard to the necessity of assumptions:
The existence of being reveals itself only in non-interpretation.
Yet, being-in-the-world presupposes that the world is interpreted in a specific way.
Any specific interpretation thus hinders access to being.
The same applies to intelligence, which, as the NFL theorems show, can be successfully applied only if assumptions are made, such as that the world exhibits regularities.
At the same time, however, these assumptions already represent an initial interpretation of the world and hinder other interpretations of the world that are not in accordance with them.

The significant similarities between Heidegger's later approach and intelligence lead to several implications that arise from phenomenology with regard to the creation of intelligence:
First, phenomenology shows that a subject is not independent of the world but is part of it and that there are close interactions; the separation between subject and world is therefore artificial and depends on the respective interpretation.
The water flea Daphnia can successfully consume food because it is in a world in which food is of such a nature that the available light stimulates the sensory system in such a way that it triggers the corresponding action.
Nevertheless, the question arises as to whether one can therefore speak of solicitations as Dreyfus and Gibson do.
Algae and their properties are indeed necessary, as is light and its properties.
However, some of Dreyfus' and Gibson's statements seem too strongly focused on the world and thus insufficiently consider the role of the agent; this, for example, when it is said that the world offers solicitations, provides guidance, and summons the agent.
Although the agent is part of the world from a phenomenological point of view, the specific characteristics of the agent, such as the degree to which it can perceive the world and gain insights from it, have to be considered as well.
This is especially because the experience of the world depends very much on the agent -- the same states of the world are perceived and, in particular, interpreted very differently by different animals, even by every human being.
The experience of the world thus also depends fundamentally on the agent itself, and it can only be understood if its active role is sufficiently taken into account.

Second, phenomenology shows that the pursuit of insight is carried out by subjects and that their presuppositions and origins must be taken into account.
This can be seen, for example, in the necessary consideration of which perceptions an agent can have, as shown in Section \ref{sec4}.
Sensory systems determine not only which aspects of the world can be perceived at all and to what degree of detail, but also how they are processed and whether they are subject to distortions, for instance.
Likewise, it is necessary to consider on which assumptions the intelligence of an agent is based; for example, in which form it is assumed that the world is subject to regularities and which other assumptions are included.

Third, the considerations in the preceding paragraph entail that humans and artificially created intelligence capture the world in fundamentally different ways:
Humans capture the world first and foremost as being-in-the-world, they experience it consciously and in a specific interpretation.
This experience is based on the specific configuration of their sensory systems and on their interpretations, which are grounded in cultures, for example.
Artificially created intelligence is also based on particular specifications, and thus interpretations, both through the sensory systems with which it is equipped and through the assumptions that are given to it.
However, the specifications and thus interpretations given to AI can be changed, while the ones for humans are relatively fixed \citep[cf.][]{975}.
Furthermore, AI is subject to far fewer specifications and interpretations than humans are.
While humans rely heavily on interpreted perceptions -- for example, a car can be experienced visually only as a car and not as a cluster of lights -- artificial systems can rely on significantly less strongly processed data \citep[cf.][pp. 40-50, 127-134]{989}.
Humans are subject to many preconceptions that they cannot question or can only question with great difficulty, as they are very strongly characterised by their interpretation.
A metaphor of \citet[p. 206]{830} illustrates this problem:
We are like sailors who have to rebuild their ship on the open sea without ever being able to dismantle it at a dock and rebuild it from scratch with the best components.

From a phenomenological perspective, AGI therefore has the advantage that it can be much closer to being, and can be much less influenced by interpretations than humans can.

\section{Meaning \& Understanding} \label{sec7}

The foregoing considerations raise the question of whether AI can be capable of assigning meaning to states of a world, i.e., to create interpretations, and if so, how it must be designed to do so.
In the following, meaning is defined as a function that is attributed to something in order to to achieve a specific goal (cf. \citealp[p. 106]{437}; \citealp[pp. 5f]{1006}).
The representation to which the function is attributed thus serves as a means for achieving a specific goal.
For example, the function of a hammer is to drive nails into walls.
Meaning is something that is attributed by an agent to something and does not exist independently of the agent.
For instance, a hammer -- or a stone -- only becomes a hammer when this function is attributed to it.
Nevertheless, the attributed function can be applied successfully only if the world in its entirety is such that the function enables the fulfilment of the goal.
For example, something can only have the function of a hammer if it is hard enough to drive a nail into the wall, there are a matching wall and nail, and the subject is able to use the item accordingly.
The successful fulfilment of an attributed function, a meaning, is therefore dependent on the world -- yet, it is not a solicitation or an offer, but a possibility.
The possibility can only be used, however, if the agent attributes it to the respective state of the world.

Closely related to meaning is understanding.
Understanding is the ability to use something in such a way that it fulfils its meaning \citep[cf.][p. 44]{768}.
For example, many people have an understanding of cars that allows them to use them as a means of transport by driving them from one place to another.
Understanding is gradual and can be more or less pronounced in terms of both efficiency and effectiveness.
For instance, some people can drive better than others and arrive at their destination faster and with less gasoline consumption.
Similarly, some people can drive in conditions in which other people can no longer drive, such as in a snowstorm or in the desert.
An agent therefore has a the greater understanding of something the more efficiently it can use it the more extensively.
As such, understanding represents a skill that describes how well something can be used in a certain functionality, i.e., with regard to an attributed meaning.

In comparison, it can be said that meaning describes the function that is attributed to something, whereas understanding describes how something has to be used to fulfil this function.
Understanding thus presupposes meaning: something can be understood only with regard to a specific meaning.
For example, a car can be understood only as a means of transport -- or as a status object or as an investment -- if the respective function is known, as each function requires a different understanding.
In the case of an investment, for instance, it is not a question of how the car can be steered with the aid of the steering wheel, but how which equipments contributes to the value of the car \citep[cf.][pp. 144f]{999}.

The definition of meaning and understanding in a functional way implies that both are present when something is successfully used to achieve a specific goal.
For example, for the water flea Daphnia, green light has the functional meaning of indicating food, and the water flea has an understanding of the light in that it uses it as an indicator of food.
The water flea also uses water as a means of transport and understands it such that it can move successfully in it.
Unlike humans, for example, the water flea does not know what light or water is from a physical point of view.
However, this is not necessary:
In the past, people also used light as an indicator of food without knowing its physical properties.
Equally, people used water for transport in the past without knowing its physical and chemical properties.
Conversely, humans today have much greater knowledge of light and water, but it is still limited -- hence there is only a difference in degree, not in kind.

It could be argued that meaning and understanding can only occur if mental states are present, which, for example, make it possible to experience them consciously.
\citet[pp. 417f]{814}, for instance, introduces the Chinese Room Argument\footnote{
The Chinese Room Argument is based on the following thought experiment:
Searle, who does not understand Chinese, is locked in a room and given three batches of Chinese characters that have no meaning to him.
In addition, Searle receives instructions in English that allow him to relate the elements of the different sets to each other to generate a fourth set, which he has to output.
Unknown to Searle, the first set is called ‘script’, the second ‘story’, the third ‘questions’, and the fourth ‘answers’.
Based on this thought experiment, Searle argues that by following the script, he can answer the questions about the story correctly and therefore, from the outside, it appears that he understands the story.
However, he does not understand the story as he does not understand the Chinese characters; instead, he only relates and manipulates these symbols according to the rules of the script.
} to argue that understanding can exist only if intentionality, which is like consciousness a mental state\footnote{
There are different views on how intentionality and consciousness are related. 
For instance, \citet[pp. 93-100]{1027} and \citet[p. 101]{983} each argue in their own way for a close relation of intentionality and consciousness.
Heidegger, on the other hand, rejects any identification of intentionality with consciousness or inner experience \citep[cf.][p. xii]{980}.
For an overview of different conceptions of intentionality, see e.g. \citet[pp. 96f]{983} and \citet{741}.
},
is present.
According to \citet[pp. 421-424]{814}, mental states can be produced only by specific physiochemical structures that have particular causal powers.
Such structures occur only in certain biological organisms: in humans, in primate species such as monkeys, and in domestic animals such as dogs.
Formal models, on the other hand, do not have the biological structures required for the causal powers and are therefore unable to constitute mental states such as intentionality and, consequently, understanding.
However, Searle does not explain why mental states can only originate from specific biological structures and how they originate from these structures.
It therefore remains unknown why other structures that can perform formal operations cannot be capable of generating intentionality as well.

It is also unclear why intentionality is necessary for understanding and what additional properties or functions intentionality, or mental states in general, contribute.
This in particular given that \citet[pp. 422-424]{814} takes a materialistic position\footnote{
This assessment is discussed controversially, as Searle does not specify how mental states arise from physiochemical structures and what kind they are.
For example, \citet[p. 291]{979} categorises Searle's approach as materialistic, whereas \citet[sect. 6]{741} denies this assessment.
} and thus does not require a separate quality from mental states that cannot arise from the material realm.
In addition, it is not clear why the conscious experience of understanding and meaning, as it occurs in humans, should be a necessary prerequisite for them.
As shown in the previous section, consciousness, in the form of being-in-the-world, enables the experience of a specific interpretation of a world, i.e., of already formed meaning and understanding.
In contrast, the generation of meaning and understanding takes place before they are accessible to consciousness in the form of experiences, their generations seems thus prior to consciousness.
It therefore seems appropriate to regard intelligence -- the creation of skills and thus of understanding as well as the attribution of meaning to phenomena -- and consciousness -- the experience of particular interpretations of a world -- as two separate aspects that are independent of each other \citep[cf.][pp. 8f]{976}.

Accordingly, in Searle's argumentation, a distinction has to be made between understanding in the functional sense and the conscious experience of understanding:
While the biological structures mentioned by Searle may be necessary for the occurrence of consciousness, understanding can occur independently of them.
With regard to the comprehension and creation of intelligence, the questions of how consciousness arises and in which agents it occurs are irrelevant.
To some extent, \citet[p. 421]{814} also advocates a functional perspective when he uses behavioural analyses to infer the existence of intentionality:
He argues that the behaviour of some animals can be explained only by attributing intentionality to them.
In this respect, Searle also advocates a functional interpretation of intentionality and understanding.

In the following, it is analysed how intelligence and, as part of it, the attribution of meaning and understanding, can be explained in a purely naturalistic way without requiring mental states such as intentionality or consciousness.
Intelligence is the ability to create a skill in which a specific state of a world, i.e., a goal, is pursued in dependence on other states of the world.
A skill hereby represents a function that leads to the fulfilment of certain states of a world.
As such, a skill represents an algorithm:
particular inputs, i.e., states of the world, lead to particular outputs, i.e., other states of the world like actions.
A simple example of a skill is the water flea Daphnia: It constitutes a skill that reacts to green light in such a way that the goal of nutrient supply is fulfilled.
A skill, i.e., the function it constitutes, is realised by an executing system.
In the case of the water flea, the executing system is the physical body, which consists of various components arranged in a specific structure:
The sensory system triggers neuronal activity in the presence of green light, which ultimately leads to swimming movements.

Since skills are functional, they can be realised in various ways and are not tied to a specific executing system.
For example, an artificial neural network for digit recognition can be realised with electrons in a silicon-based computer chip, as well as with optical waves in a nanophotonic medium \citep{835}.
Nevertheless, a skill is only realised due to an executing system: the executing system is therefore necessary and constitutive.
Accordingly, the executing system is material and skill at once:
The skill results from the properties of the executing system, such as its structure \citep[cf.][pp. 202-206]{843}.

The skill of the water flea Daphnia is functionally a very simple skill.
Intelligence is functionally more complex but subject to the same considerations as it is also a skill.
In functional terms, intelligence is an optimisation algorithm whose goal is to develop a skill that achieves a specific goal under specific circumstances.
An example of an optimisation algorithm is a reinforcement learning algorithm.

The attribution of meaning results from the creation of a suitable skill.
For example, a reinforcement learning algorithm can create a skill that is optimised based on the reward for fulfilling the goal of nutrient supply:
If the skill created leads to movement towards green light, as this turns out to be beneficial for the achievement of the goal, green light takes on the meaning of serving as a means of nutrient supply.
Equally, a skill implies an understanding of something if it is successfully used to achieve the goal.
The skill above, for example, implies an understanding of green light if it can be successfully utilised to achieve the goal of nutrient supply.

Based on these considerations, it is possible to define information:
Information consists of representations that are used functionally.
For example, a representation that indicates the presence of green light becomes information in that it is used by the water flea to fulfil the goal of nutrient supply.
From a functional point of view, a representation is information about the state of a world on which the representation is dependent \citep[cf.][pp. 300f]{979}.
Like skills, information is functional and therefore also not bound to a specific executing system, a specific medium, but can be realised in different ways.
As with skills, however, media are necessary and constitutive.
Similarly, the ability of a medium to provide information results from and depends on its properties:
A medium, for example an electron released during neuronal activity, can be information about a state of the world exactly then, if it represents it.

\citet[pp. 121-123]{983} argue that representations cannot serve as a basis for understanding:
To know that a representation corresponds to the represented, one must first grasp the represented directly, i.e., non-representationally -- but this is not possible from a representational point of view, since one can only grasp representations of something but never the represented itself as it is.
However, as the preceding considerations here and in Section \ref{sec4} show, representations are not depictions that have to be created on the basis of what is to be represented, but they are inherently dependent on that.
In consequence, representations may be incomplete in the way that they reflect only partially the states they represent, but they are grounded in them and thus correspond to them; an assignment is therefore not necessary \citep[cf.][p. 402]{743}.

Overall, this naturalistic approach makes it possible to explain the realisation of intelligence without having to resort to controversially discussed and ambiguous concepts such as cognition, mind, thought, or intentionality.
The approach also dispenses with the need for consciousness, which is considered something that can co-occur with intelligence but is functionally independent of it.
The approach advocated here does not take a position on how consciousness arises and in which agents it is present; whether, for example, insects such as the water flea Daphnia, the primate species and domestic animals mentioned by Searle -- or certain forms of AI -- exhibit consciousness and what its nature is.
The approach also allows one to avoid several controversial assumptions, such as that a world offers solicitations or that agents are summoned by the world.
Moreover, the approach makes it possible to solve the relevance problem:
Meaning results from a representation taking on a particular function within a skill.
The function is attributed by intelligence by drawing a relation between the represented state of the world to which the function is attributed and the state of the world to be achieved, i.e., the goal.

\section{World Model \& Reasoning} \label{sec8}

The entirety of all knowledge, i.e., all skills as well as all non-goal-orientated knowledge, such as knowledge about particular states of a world, is often referred to as world model in the field of AI.
Intelligence, i.e., the creation of skills, is thus about the expansion of a world model.
Various methods are available for this purpose, which are analysed in the following.

World models can differentiate from each other with respect to their complexity, for example, with regard to the amount of knowledge they include, but also whether they take the temporal dimension into account.
The Daphnia water flea represents a very simple world model in which the temporal dimension is not taken into account and which is mainly composed of simple, action-orientated knowledge, such as that it is helpful for the goal of nutrient supply to swim towards green light.\footnote{
To a certain extent, the temporal dimension is accounted for insofar as, for example, circadian rhythms influence the behaviour of the water flea.
However, the temporal dimension is not incorporated in such a way that future states or future actions are considered.
}
In comparison, humans have a very complex world model that takes into account the past and the future, and describes many states of the world in detail.\footnote{
Despite its greater complexity, the human world model, like that of the water flea, is in principle action-orientated.
This is illustrated by the environmental dependency syndrome, which can result from focal unilateral frontal lobe lesions and causes people to react directly and compulsively to environmental stimuli, for example, when they see a bed, they lie down \citep{836}.}
Similarly, in the world model of the water flea, few states of the world are attributed only few meanings; algae, for example, serve only as food.
In human world models, on the other hand, many states of the world are attributed many meanings. For example, plants are used as food, medicine, wrapping material, decoration, poison, and combustion material.

Knowledge that forms a world model can originate from three types of sources:
Prior knowledge refers to all knowledge made available to an agent, for example, in the form of assumptions that serve as the basis for intelligence.\footnote{
For an overview of prior knowledge in humans, often called core knowledge, see \citet{975}.
For a possible implementation of core knowledge in the field of AI, see \citet[pp. 4, 9-12]{282}.
}
Perceived knowledge refers to all knowledge an agent gains through perception, for example, by vision.
Derived knowledge refers to all knowledge an agent derives from other knowledge, for example, by inferential methods such as deduction and induction.

Intelligence is about the derivation of knowledge with the aim of determining actions that allow the fulfilment of given goals:
Skills are created by deriving them from already present and possibly perceived knowledge, e.g., by observing new unknown states and expanding existing skills accordingly \citep[cf.][ch. 4-7]{829}.
Intelligence can occur only through the derivation of knowledge:
If skills were provided in the form of prior knowledge, they would not be created and therefore would not meet the definition of intelligence outlined in Section \ref{sec2}.
Also, no new skills can be gained purely from perception, as perception has to be interpreted and set in relation to the goals to be achieved in order to become skills.

The derivation of new knowledge from existing knowledge is achieved by reasoning.
Reasoning comprises various methods that make it possible to draw more or less reliable conclusions from existing knowledge.
Among others, reasoning includes the three inference methods deduction, induction, and abduction.
Deduction allows to derive certain conclusions, i.e., the truth of a conclusion necessarily follows from the truth of the premises.
For example, if swans are birds and all birds lay eggs, then swans lay eggs.
Induction allows generalisations, i.e., to make predictions about hitherto unknown states of a world, but is uncertain.
For example, if all the swans one has seen are white, one can infer that all swans that exist are white.
Abduction allows to infer from a known state of a world to another state of the world that implies the known one \citep{711}.
For example, from wet grass one can infer that it has rained.
Abduction is in general an uncertain conclusion, since there can also be other implying states of the world, for example, a lawn sprinkler.

In addition to that, abduction is the most powerful inference method as it allows the introduction of new, composed concepts:
For example, one can infer abductively from the observation of apples falling from trees the concept of gravity.\footnote{
Concepts are -- from the perspective of the approach presented in this article -- synonymous with representations; the two terms differ primarily in that the term concept is common in the field of philosophy and psychology, whereas the term representation is primarily used in the field of AI.
In Section \ref{sec5}, representations are defined as states that are dependent on one or more other states.
This implies that representations can represent states of a world that do not themselves constitute representations, as well as states that themselves constitute representations.
Representations therefore also include inferred representations.
Some approaches consider as concepts only representations that do not just originate from direct perception but also include theoretical features \citep[cf.][pp. 4-8]{1004}.
This demarcation is ambiguous, however, and cannot be based on a qualitative difference:
All sensory perceptions are already theoretical in nature due to the way they are perceived, as well as the way they are processed in the sensory systems (cf. sect. \ref{sec4}; \citealp[EP 2 p. 227]{1023}).
Another possible definition is based on the assumption that representations are only concepts if they are used to explain data but cannot be perceived themselves \citep[pp. 14f]{738}.
Here too, however, it is not possible to make a clear distinction; bacteria and electrons were originally purely theoretical concepts, but can now be perceived with the aid of microscopes.
}
Although intelligence relies only on existing knowledge and perceptions, it is thus possible, with the help of abduction, to extend world models and to create new representations (\citealp[sect. 5]{711}; \citealp{754}).
Nevertheless, the creation of something new is limited insofar as everything new has to be based on something known; all newly formed representations originate from existing representations (cf. \citealp[bk. II ch. I par. 24]{1028}; \citealp[p. 193]{834}).
For example, from the two existing representations of a red line and a green circle, it is possible to create a new representation of a red circle.
However, it is not possible to create a representation of a new colour or shape without drawing on other existing representations.

It is unclear to what extent humans use the three inference methods deduction, induction and abduction and to what degree they use other, additional reasoning methods.
For example, instead of induction, Bayes' theorem could also be applied \citep{661}.\footnote{
Although it is unclear exactly which methods are used by humans and animals, methods for recognising regularities are widely used, as a study of \citet{844} on pigeons illustrates.
}
Humans also appear to possess the ability for causal reasoning as prior knowledge \citep{291}, although this could be derived inductively as well.
Further research is therefore needed on which reasoning methods should serve as a basis for intelligence, particularly with regard to the development of AGI.
The reasoning processes on which humans rely have proven to be advantageous in evolutionary terms and could therefore be viewed positively.
Nevertheless, evolutionary development is a continuous process and is dependent on the environment and human limitations, such as the performance of the brain.
Furthermore, it seems to be evolutionarily advantageous if humans are equipped with as many skills as possible right from the start; for example, it is easier to recognise causality if it is already known and does not have to be derived using intelligence.
With respect to the development of AGI, however, it can be advantageous to provide it with only the most foundational methods possible as prior knowledge in order to minimise the number of potentially incorrect assumptions.
Although, as shown in Section \ref{sec2}, it is necessary to specify assumptions for intelligence due to the NFL theorems, these should be determined prudently and be as foundational as possible.
In this respect, further research is also needed on whether there is one set of assumptions which represents an optimum for AGI -- or whether it may be advantageous to create several instances of AGI with different sets of assumptions to achieve greater variety, and thereby more powerful forms of intelligence.

In addition to the aforementioned methods of reasoning, at least two further methods are required for processing representations, and which, for example, provide the foundation for the formation of new, derived representations: abstraction and classification.
Abstraction describes the ability to select specific features of a representation and to disregard all other features.
For example, from the representation of a green circle or a green tree, the abstraction green can be created.
Abstraction, like abduction, thus allows the introduction of new representations.
However, abstraction can only form new representations by removing features from existing representations.
Abduction, on the other hand, can form new representations by combining different features from various representations and is therefore more powerful, as it can create composed representations.
Abstraction occurs extensively in sensory systems in biological organisms, as it makes it possible to significantly reduce the number of representations in order to represent only relevant states of a world \citep[cf.][pp. 66f]{1006}.
Beyond this, abstraction is seen as an important method for creativity \citep{837}.

Classification is a method that is applied in two ways.
First, classification can be based on abstraction:
Features that have been abstracted can serve as a basis for a common classification of different representations.
For example, the abstraction green allows all representations that contain this feature to be grouped together, e.g., green circles together with green trees.
Second, classification takes place to form individual representations from the temporally continuous stream of perception.
This often takes place in sensory systems and depends on their design.
For instance, the configuration of neurons and their firing rate determine whether several flashes of light are classified into several separate or one combined representation.
Classification is therefore one of the most elementary methods used to process representations.

Reasoning methods and other methods, such as abstraction and classification, for processing representations and deriving new knowledge hence represent methods that can be applied by intelligence to develop a world model -- and thus skills.
At least some of the methods have to be provided for intelligence and constitute assumptions on which it is based.
The methods are hence a concretisation of the necessary assumptions discussed in Section \ref{sec3}, which have to be provided to intelligence due to the NFL theorems.
The conclusions drawn via these methods are reliable only if the methods fit the respective world for which they are assumed.
For example, induction can be successfully applied only if the world is such that generalisation leads to success \citep[cf.][sect. IV]{1033}.

\section{Viability \& Construction} \label{sec9}

The value of a world model is judged by its functional performance:
The more extensively goals can be achieved with the help of skills that are part of the world model, the more useful the model is (\citealp[pp. 116-128]{977}; \citealp[pp. 136f]{989}).
\citet[pp. 14, 68f]{977} uses the term viability to describe how successful an agent is in achieving its goals.
Accordingly, the aim in developing a world model is to ensure that it corresponds to a world as far as possible, i.e., that the model is compatible with the world in the sense that the chosen actions lead to the fulfilment of the goals.
In contrast, the world model is not meant to depict the truth, i.e., the world as it is \citep[pp. 109-114]{977}.

Eliminating the need for truth and instead aligning a world model solely on viability offers several advantages.
First, this allows complex issues to be represented in simplified forms, as long as they are functionally precise enough.
For example, humans use simplified descriptions of how objects move, e.g., to predict the trajectory of a throw, which ignore many factors and are not true but functional \citep[p. 10]{282}.
Second, it is not necessary that truth has to be perceived.
As shown in Section \ref{sec4}, humans only experience representations of states of the world through their sensory systems but not the states themselves \citep[cf.][pp. 312f]{831}.
Equally, as shown in Section \ref{sec6}, human conscious experience does not allow direct access to phenomena, i.e., truth, but only offers an interpretation, i.e., an experience based on a world model \citep[cf.][pp. 317f]{831}.
It is therefore unclear on what foundations a world model based on truth can be developed.
The approach advocated here does not exclude the possibility of perceiving truth but does not require it either, which makes the approach less presuppositional.

Even though the approach does not aim to reflect the truth of a world, it nevertheless assumes a correspondence with the world:
Although perceptions are indirect and only provide representations and not the world itself, the representations are dependent on the world and therefore correspond to it.
For example, an opsin sends a particular neural signal, a representation, precisely when it perceives light.
The neural signal is only a representation of the light and not the light itself, but it depends on it.
In this way, world models are grounded in the world and correspond to it.
The at least partial correspondence of the world model with the world is shown by its functional success: without correspondence, a world model could not be viable.

Applying reasoning methods to the perceived correspondences allows a multitude of different conclusions to be drawn.
The reason for this is that some of the reasoning methods are uncertain and contingent and therefore allow only possible but uncertain conclusions, i.e., hypotheses.
Not all hypotheses can be directly, e.g., empirically, assessed, and some hypotheses may imply the same perceivable correspondences.
For example, the observation of apples falling from trees can be explained by the hypothesis that gravitational forces act on them.
Alternatively, the observation can also be explained as an effect of spacetime curvature without the apples being exposed to forces.
Both hypotheses represent viable conclusions, and without further observations and reasoning, neither can be proven to be superior, i.e., more viable, than the other \citep[pp. 113f]{977}.
Consequently, as long as the conclusions drawn are viable, any of the most different conclusions can be accepted, each being as valid as any other.
\citet[p. 118]{977} describes this constructivist position, in which any representation can be constructed as long as it is functional and corresponds to a world, as follows:
"What we ordinarily call reality is the domain of the relatively durable perceptual and conceptual structures which we manage to establish, use, and maintain in the flow of our actual experience."

Hypotheses can be evaluated using explanatory virtues to determine which of several competing hypotheses should be preferred.
For example, hypotheses can be preferred that are simpler or make more comprehensive statements \citep[CP 6.447]{1022}.
However, the significance of explanatory virtues is unclear and it is not clear to what extent they enable an assessment of hypotheses (cf. \citealp [pp. 6f]{711}; \citealp[sect. 3]{487}).
Furthermore, it is unclear what their assessments are based on.
On the one hand, it is conceivable that they are rooted in a statement about the nature of a world, such as that the world is simpler rather than more complex.
This, however, is an assumption and it may not apply to the world in which the virtue is used; with the consequence that conclusions derived from it may also not apply to that world.
On the other hand, virtues can be derived from existing assumptions -- for example, that the preferability of a hypothesis is measured by its functionality.
However, with this, virtues do not offer additional assessment opportunities beyond the existing ones.

The contingency in the processing of representations, i.e., the possibility of drawing not only one conclusion but a multitude of different ones, applies not only to inference methods but also to classification:
Here, too, it is necessary to carry out the classification on the basis of specific assumptions, i.e., virtues, which determine, for example, how many classes should be created, or on the basis of which criteria elements should be classified as similar or different to each other.\footnote{
An illustration of this is provided by the Chinese Restaurant Process, which utilises the virtue simplicity to determine whether an element should be assigned to an existing class or whether a new class should be created for it \citep[p. 1284]{206}.
}

The representational character of intelligence, which is due to the indirectness of perception, constitutes its potency \citep[cf.][pp. 400f]{754}:
From existing representations, new representations can be formed that are grounded in a world but that do not necessarily correspond to it completely; in other words, it is possible to form representations that deviate from the world.
This enables the formation of constructs and the realisation of planning, i.e., the creation of what-if scenarios and the prediction of what future states of a world will be like without these states actually existing.

Overall, the considerations thus show that world models are not images of a world that are becoming increasingly detailed and depict ever more aspects of the world.
Instead, world models are collections of contingent and uncertain conclusions that aim for the greatest possible correspondence with a world and the greatest possible viability, i.e., the possibility of achieving goals.
World models may seem like truth from human conscious experience, since one is accustomed to them and since they correspond to the world, but nevertheless they are only constructs, as \citet[p. 314]{831} metaphorically describes:
So what is truth? A mobile legion of metaphors, metonymies, anthropomorphisms, in short a sum of human relations that have been poetically and rhetorically intensified, transferred, adorned, and which, after long use, seem firm, canonical and binding to a people: truths are illusions of which one has forgotten that they are such.

\section{Agentness \& Interrelation} \label{sec10}

Intelligence is applied to develop a world model that enables the fulfilment of goals as comprehensively as possible.
In this way, the world model should have the greatest possible correspondence with the world and make it possible to identify actions that allow to influence the world in such a way that the goals are achieved as far as possible.
By world is meant everything that is; in phenomenology this is referred to as the totality of phenomena.
However, agents, including humans, animals and AGI, cannot access the world in which they are situated in its entirety and in a direct way, as has been shown in the discussion of phenomenology in Section \ref{sec6}.
Instead, agents are faced with the challenge that they can grasp only corresponding representations of a world through perception, which usually concern only a small part of the world and can be distorted.

The correspondences usually appear to be in the form of a temporally continuous stream of perception.\footnote{
This applies at least to the world in which humans live and to many worlds created by humans, such as computer games.
There are also worlds that do not involve a temporal aspect, such as some logical puzzles like Sudoku.
Mathematics is in general also not based on temporal aspects, although it can be used to represent them.
}
Nevertheless, it is not clear to what degree time (and space) reflect a basic constitution of the world and to what extent it is only an interpretation, i.e., the experience of a world model \citep[cf.][pp. 78-80]{1029}.
The basis and starting point of all applications of intelligence is thus, as Heidegger pointed out, this subjective perspective of perception on the basis of which the world has to be functionally comprehended.

The assumption that there is a world which can be perceived often proves to be helpful from a functional point of view.
The approach advocated here only assumes the existence of such a world, but makes no further specific assumptions, such as that the world is material; it is only necessary that correlations can be perceived.\footnote{
Consequently, it makes no difference whether an agent perceives the representations reflecting the correspondence directly from a world or by means of a computer, as is the case in the brain in a vat thought experiment of \citet[ch. 1]{1030}.
}
Equally, the division of a world into a self, in the functional sense, and an environment is often helpful from an agent's perspective, whereby the division is based on functional criteria \citep[cf.][]{842}.
For example, it can be helpful to differentiate between one's own body, i.e., the executing system, and the rest of the world, since the own body forms a spatial and temporal unit, can be perceived differently, and manipulations to the body lead to different effects compared to the rest of the world \citep[cf.][pp. 325-328]{1006}.

Irrespective of the functional advantages of dividing a world into a self and an environment, an agent is in general part of the world and as such is subject to close interactions with the rest of the world in many ways.
For example, the possibilities of the agent's perception are determined by the world.
The perception of light requires not only the presence of light, but also many other aspects of the world influence it.
For instance, light propagates much better in air than in water, which is why visibility on land is much better than in water.
\citet{838} argue that as a result, the migration of animals from water to land about 400 million years ago led to a significant improvement in the eyes and, consequently, to more elaborate behaviour, as the more extensive perceptual possibilities allowed for more sophisticated planning.
An agent is therefore not just something that perceives a world but is formed by it.
Adaptations between agents and their environment take place in many respects, as they are functionally advantageous (cf. \citealp[pp. 114f, 221-223, 228]{1006}; \citealp[p. 128]{989}).

One advantage of adaptations is that they make it possible to minimise the need to process representations in order to identify optimal actions.
As an example, the sensory system for sound waves of female crickets is connected to their locomotor system in such a way that melodies produced by male crickets automatically lead to movements in the direction of their location, whereas all other sounds do not cause a reaction \citep[pp. 1091-1093]{839}.
A comprehensive analysis of all the sounds heard and filtering out melodies, as occurs to some extent in humans, can therefore be avoided.
Another example is monkeys whose colour receptors are adapted to the colours of nutritious fruits, enabling them to recognise the fruits more easily and thus reducing the demands on perceptual analysis \citep[p. 128]{989}.
The adaptation of agents to their environments to achieve more efficient and effective goal fulfilment plays an important role in robotics \citep[p. 19]{976}.
Brooks' robot Herbert, which was analysed in Section \ref{sec5}, was significantly more efficient than other models developed at that time, as it was strongly optimised to fulfil the objectives with minimal use of resources.
For example, image analysis for object detection was avoided by using simpler but viable ultrasonic sensors instead.

Such adaptations of agents are particularly helpful in regard to executing skills, i.e., when particular actions have to be performed because of particular states of a world.
Adaptions to support intelligence are much more difficult because states of a world have not yet been attributed a specific meaning, and it is therefore unknown which states of the world should be used in which way under what circumstances.
Nevertheless, from a functional point of view, it seems advisable to design intelligent agents in such a way that they are able to manipulate a world as comprehensively as possible.
On the one hand, this gives them more options for action and enables them to identify more favourable actions to achieve their goals.
On the other hand, manipulating a world allows hypotheses to be tested and falsified, which allows world models to be developed with greater correspondence and thus greater viability.
\citet[pp. 80, 86f, 117]{998} discusses the thesis of whether the nature of an agent's embodiment constrains or determines the concepts it can acquire, arguing that if the thesis "is correct, then human beings could not share thoughts with differently embodied aliens because they could not possess the same concepts".
The nature of embodiment indeed influences the concepts that can be created, e.g., through the possibilities and limitations of perception, as well as the reasoning methods that can be applied.
However, this does not imply that it is impossible for agents with different embodiments to have the same concepts.
For example, a car may be perceived in different ways and the corresponding concepts may be created by different reasoning methods, but the created concepts may still represent the same states of the world and be functionally the same.
Consequently, the development of different concepts due to different conditions and contingencies is possible but not inevitable.
In addition, agents can synchronise their concepts through communication, as is common between humans; in this article, for example, through definitions and deliberations.

\section{Conclusion} \label{sec11}

The aim of the article is to identify and analyse principles that have to be considered for the creation of AGI.
Based on the analyses in the preceding sections, the following findings are drawn:
The purpose of AGI is the fulfilment of given goals in a partially unknown world.
To achieve these goals, AGI must develop skills, i.e., instructions for action that enable the fulfilment of the goals depending on states of the world.
Novel skills for hitherto unknown conditions can be created by intelligence, which is based on the application of various reasoning methods such as deduction, induction and abduction, as well as other methods such as abstraction and classification.
Due to the nature of perception, intelligence cannot grasp a world as it is but can only use representations that reflect the world indirectly and possibly incompletely and distorted.
As representations correspond to the world, intelligence can draw conclusions from them about the world using uncertain and contingent reasoning methods.
This makes it possible to attribute functions to representations as to how they can be used to achieve goals; by doing so, representations are attributed meaning.
The totality of all existing knowledge forms a world model, which contains, for example, all skills and which can be expanded with the help of reasoning methods and new perceptions.
The value of a world model is functionally determined by its viability, i.e., its potential to fulfil the goals.
Due to the uncertainty and contingency of the reasoning methods, many different possible viable conclusions can be drawn.
As a consequence, the world model is constructivist, i.e., the conclusions drawn do not represent the world truthfully but only correspond to it.
The methods of reasoning represent assumptions about the world; due to the NFL theorems, it is necessary to provide at least some assumptions as a basis for intelligence.
However, intelligence is only successful if the assumptions apply to the world in which it is used, which is why they should be determined prudently.
Overall, intelligence is considered an algorithm for an optimisation problem whose task is to find optimal actions to achieve particular goals in a partially unknown world.
This interpretation relies on a naturalistic approach and does not require the assumption of mental features, such as consciousness, which are considered to be independent of intelligence.
The performance of AGI is determined by how comprehensively it can perceive the world, how comprehensively it can manipulate the world, how comprehensively it can apply reasoning and other methods, and how efficient and consistent with the world the assumptions on which it is based are.

The considerations presented in this article also represent a constructivist-generated world model, a specific interpretation of all that is.
From the author's point of view, based on conscious experiences, cultural influences, knowledge given at birth, and conclusions based on these, the considerations presented here appear to have the highest achievable correspondence with the world.
Whether these considerations are viable, i.e., functional, and offer the possibility of creating AGI has yet to be determined.
The considerations made here offer a new perspective on the development of AGI insofar as, in contrast to numerous other approaches, they focus away from the utilisation of knowledge towards the generation of knowledge by means of reasoning methods, in particular deduction, induction, and abduction.
Abduction is a method that has so far received relatively less attention in the field of AI, but also in the field of philosophy; at the same time, it is the most powerful inference method, as it allows the generation of new, composed representations.
Consequently, abduction, as well as other topics addressed in this article, requires a more detailed examination and further research.

The considerations developed in the article also allow for various considerations regarding generative AI approaches, which are currently gaining ground, particularly in the form of large language models and large multimodal models, and which are considered by many to be the closest to AGI currently available.
While these models deliver results that are considered impressive by many, they are based on an enormous amount of training data.
They seem to be able to apply reasoning methods and solve unfamiliar problems, but only to a limited extent.
From the perspective of the conception of intelligence developed here, these models are primarily, but not exclusively, based on skills rather than intelligence \citep{841}.
The models also have the disadvantage that the knowledge provided to them does not represent very few fundamental assumptions about the world but an already highly processed and very specific interpretation of the world from a human perspective.
As a result, the models are founded on representations similar to those of humans, which makes communication much easier, but the models cannot develop their own, possibly more viable representations.

Yet, this is precisely where the opportunity of AGI could lie, especially from a philosophical point of view, but also from a scientific point of view:
AGI can receive much more raw and comprehensive representations of the world compared to humans and process them by other means, which, to draw on Neuropath's metaphor, can allow it to build a new ship from scratch at a dock using better components.
This new ship, an almost new interpretation of the world, could represent a comprehensive enrichment for humanity.

\newpage

\bibliography{sn-bibliography}

\begin{thebibliography}{}
\renewcommand{\doi}[1]{\url{https://doi.org/#1}}
\bibcommenthead

\bibitem [\protect \citeauthoryear {%
Abramson%
\ \protect \BOthers {.}}{%
Abramson%
\ \protect \BOthers {.}}{%
{\protect \APACyear {2024}}%
}]{%
789}
\APACinsertmetastar {%
789}%
\begin{APACrefauthors}%
Abramson, J.%
, Adler, J.%
, Dunger, J.%
, Evans, R.%
, Green, T.%
, Pritzel, A.%
\BDBL {}others%
\end{APACrefauthors}%
\unskip\
\newblock
\APACrefYearMonthDay{2024}{}{}.
\newblock
{\BBOQ}\APACrefatitle {Accurate structure prediction of biomolecular interactions with AlphaFold 3} {Accurate structure prediction of biomolecular interactions with alphafold 3}.{\BBCQ}
\newblock
\APACjournalVolNumPages{Nature}{}{}{1--3}
\newblock

\newblock

\PrintBackRefs{\CurrentBib}

\bibitem [\protect \citeauthoryear {%
Baker%
, Modrell%
\BCBL {}\ \BBA {} Gillis%
}{%
Baker%
\ \protect \BOthers {.}}{%
{\protect \APACyear {2013}}%
}]{%
811}
\APACinsertmetastar {%
811}%
\begin{APACrefauthors}%
Baker, C.V.%
, Modrell, M.S.%
\BCBL {} Gillis, J.A.%
\end{APACrefauthors}%
\unskip\
\newblock
\APACrefYearMonthDay{2013}{}{}.
\newblock
{\BBOQ}\APACrefatitle {The evolution and development of vertebrate lateral line electroreceptors} {The evolution and development of vertebrate lateral line electroreceptors}.{\BBCQ}
\newblock
\APACjournalVolNumPages{Journal of Experimental Biology}{216}{13}{2515--2522}
\newblock

\newblock

\PrintBackRefs{\CurrentBib}

\bibitem [\protect \citeauthoryear {%
Beckmann%
, K{\"o}stner%
\BCBL {}\ \BBA {} Hip{\'o}lito%
}{%
Beckmann%
\ \protect \BOthers {.}}{%
{\protect \APACyear {2023}}%
}]{%
743}
\APACinsertmetastar {%
743}%
\begin{APACrefauthors}%
Beckmann, P.%
, K{\"o}stner, G.%
\BCBL {} Hip{\'o}lito, I.%
\end{APACrefauthors}%
\unskip\
\newblock
\APACrefYearMonthDay{2023}{}{}.
\newblock
{\BBOQ}\APACrefatitle {An alternative to cognitivism: computational phenomenology for deep learning} {An alternative to cognitivism: computational phenomenology for deep learning}.{\BBCQ}
\newblock
\APACjournalVolNumPages{Minds and Machines}{33}{3}{397--427}
\newblock

\newblock

\PrintBackRefs{\CurrentBib}

\bibitem [\protect \citeauthoryear {%
Berglund%
\ \protect \BOthers {.}}{%
Berglund%
\ \protect \BOthers {.}}{%
{\protect \APACyear {2023}}%
}]{%
787}
\APACinsertmetastar {%
787}%
\begin{APACrefauthors}%
Berglund, L.%
, Tong, M.%
, Kaufmann, M.%
, Balesni, M.%
, Stickland, A.C.%
, Korbak, T.%
\BCBL {} Evans, O.%
\end{APACrefauthors}%
\unskip\
\newblock
\APACrefYearMonthDay{2023}{}{}.
\newblock
{\BBOQ}\APACrefatitle {The reversal curse: Llms trained on" a is b" fail to learn" b is a"} {The reversal curse: Llms trained on" a is b" fail to learn" b is a"}.{\BBCQ}
\newblock
\APACjournalVolNumPages{arXiv preprint arXiv:2309.12288}{}{}{}
\newblock

\newblock

\PrintBackRefs{\CurrentBib}

\bibitem [\protect \citeauthoryear {%
Brooks%
}{%
Brooks%
}{%
{\protect \APACyear {1991}}%
{\protect \APACexlab {{\protect \BCnt {1}}}}}]{%
751}
\APACinsertmetastar {%
751}%
\begin{APACrefauthors}%
Brooks, R.A.%
\end{APACrefauthors}%
\unskip\
\newblock
\APACrefYearMonthDay{1991{\protect \BCnt {1}}}{}{}.
\newblock
{\BBOQ}\APACrefatitle {Intelligence Without Reason} {Intelligence without reason}.{\BBCQ}
\newblock
\APACjournalVolNumPages{MIT AI Laboratory: AI Memos}{}{}{}
\newblock

\newblock

\PrintBackRefs{\CurrentBib}

\bibitem [\protect \citeauthoryear {%
Brooks%
}{%
Brooks%
}{%
{\protect \APACyear {1991}}%
{\protect \APACexlab {{\protect \BCnt {2}}}}}]{%
734}
\APACinsertmetastar {%
734}%
\begin{APACrefauthors}%
Brooks, R.A.%
\end{APACrefauthors}%
\unskip\
\newblock
\APACrefYearMonthDay{1991{\protect \BCnt {2}}}{}{}.
\newblock
{\BBOQ}\APACrefatitle {Intelligence without representation} {Intelligence without representation}.{\BBCQ}
\newblock
\APACjournalVolNumPages{Artificial intelligence}{47}{1-3}{139--159}
\newblock

\newblock

\PrintBackRefs{\CurrentBib}

\bibitem [\protect \citeauthoryear {%
Cabrera%
}{%
Cabrera%
}{%
{\protect \APACyear {2017}}%
}]{%
487}
\APACinsertmetastar {%
487}%
\begin{APACrefauthors}%
Cabrera, F.%
\end{APACrefauthors}%
\unskip\
\newblock
\APACrefYearMonthDay{2017}{}{}.
\newblock
{\BBOQ}\APACrefatitle {Can there be a Bayesian explanationism? On the prospects of a productive partnership} {Can there be a bayesian explanationism? on the prospects of a productive partnership}.{\BBCQ}
\newblock
\APACjournalVolNumPages{Synthese}{194}{4}{1245--1272}
\newblock

\newblock

\PrintBackRefs{\CurrentBib}

\bibitem [\protect \citeauthoryear {%
Carey%
}{%
Carey%
}{%
{\protect \APACyear {2000}}%
}]{%
1004}
\APACinsertmetastar {%
1004}%
\begin{APACrefauthors}%
Carey, S.%
\end{APACrefauthors}%
\unskip\
\newblock
\APACrefYearMonthDay{2000}{}{}.
\newblock
{\BBOQ}\APACrefatitle {The origin of concepts} {The origin of concepts}.{\BBCQ}
\newblock
\APACjournalVolNumPages{Journal of Cognition and Development}{1}{1}{37--41}
\newblock

\newblock

\PrintBackRefs{\CurrentBib}

\bibitem [\protect \citeauthoryear {%
Chollet%
}{%
Chollet%
}{%
{\protect \APACyear {2019}}%
}]{%
674}
\APACinsertmetastar {%
674}%
\begin{APACrefauthors}%
Chollet, F.%
\end{APACrefauthors}%
\unskip\
\newblock
\APACrefYearMonthDay{2019}{}{}.
\newblock
{\BBOQ}\APACrefatitle {On the measure of intelligence} {On the measure of intelligence}.{\BBCQ}
\newblock
\APACjournalVolNumPages{arXiv preprint arXiv:1911.01547}{}{}{}
\newblock

\newblock

\PrintBackRefs{\CurrentBib}

\bibitem [\protect \citeauthoryear {%
Cummings%
\ \BBA {} Bauchwitz%
}{%
Cummings%
\ \BBA {} Bauchwitz%
}{%
{\protect \APACyear {2024}}%
}]{%
793}
\APACinsertmetastar {%
793}%
\begin{APACrefauthors}%
Cummings, M.L.%
\BCBT {}\ \BBA {} Bauchwitz, B.%
\end{APACrefauthors}%
\unskip\
\newblock
\APACrefYearMonthDay{2024}{}{}.
\newblock
{\BBOQ}\APACrefatitle {Unreliable Pedestrian Detection and Driver Alerting in Intelligent Vehicles} {Unreliable pedestrian detection and driver alerting in intelligent vehicles}.{\BBCQ}
\newblock
\APACjournalVolNumPages{IEEE Transactions on Intelligent Vehicles}{}{}{}
\newblock

\newblock

\PrintBackRefs{\CurrentBib}

\bibitem [\protect \citeauthoryear {%
Dohare%
\ \protect \BOthers {.}}{%
Dohare%
\ \protect \BOthers {.}}{%
{\protect \APACyear {2024}}%
}]{%
794}
\APACinsertmetastar {%
794}%
\begin{APACrefauthors}%
Dohare, S.%
, Hernandez-Garcia, J.F.%
, Lan, Q.%
, Rahman, P.%
, Mahmood, A.R.%
\BCBL {} Sutton, R.S.%
\end{APACrefauthors}%
\unskip\
\newblock
\APACrefYearMonthDay{2024}{}{}.
\newblock
{\BBOQ}\APACrefatitle {Loss of plasticity in deep continual learning} {Loss of plasticity in deep continual learning}.{\BBCQ}
\newblock
\APACjournalVolNumPages{Nature}{632}{8026}{768--774}
\newblock

\newblock

\PrintBackRefs{\CurrentBib}

\bibitem [\protect \citeauthoryear {%
Dreyfus%
}{%
Dreyfus%
}{%
{\protect \APACyear {2002}}%
}]{%
776}
\APACinsertmetastar {%
776}%
\begin{APACrefauthors}%
Dreyfus, H.L.%
\end{APACrefauthors}%
\unskip\
\newblock
\APACrefYearMonthDay{2002}{}{}.
\newblock
{\BBOQ}\APACrefatitle {Intelligence without representation--Merleau-Ponty's critique of mental representation The relevance of phenomenology to scientific explanation} {Intelligence without representation--merleau-ponty's critique of mental representation the relevance of phenomenology to scientific explanation}.{\BBCQ}
\newblock
\APACjournalVolNumPages{Phenomenology and the cognitive sciences}{1}{4}{367--383}
\newblock

\newblock

\PrintBackRefs{\CurrentBib}

\bibitem [\protect \citeauthoryear {%
Dreyfus%
}{%
Dreyfus%
}{%
{\protect \APACyear {2007}}%
}]{%
701}
\APACinsertmetastar {%
701}%
\begin{APACrefauthors}%
Dreyfus, H.L.%
\end{APACrefauthors}%
\unskip\
\newblock
\APACrefYearMonthDay{2007}{}{}.
\newblock
{\BBOQ}\APACrefatitle {Why Heideggerian AI failed and how fixing it would require making it more Heideggerian} {Why heideggerian ai failed and how fixing it would require making it more heideggerian}.{\BBCQ}
\newblock
\APACjournalVolNumPages{Philosophical psychology}{20}{2}{247--268}
\newblock

\newblock

\PrintBackRefs{\CurrentBib}

\bibitem [\protect \citeauthoryear {%
Dreyfus%
\ \BBA {} Dreyfus%
}{%
Dreyfus%
\ \BBA {} Dreyfus%
}{%
{\protect \APACyear {1986}}%
}]{%
1011}
\APACinsertmetastar {%
1011}%
\begin{APACrefauthors}%
Dreyfus, H.L.%
\BCBT {}\ \BBA {} Dreyfus, S.E.%
\end{APACrefauthors}%
\unskip\
\newblock
\APACrefYear{1986}.
\newblock
\APACrefbtitle {Mind over machine} {Mind over machine}.
\newblock
\APACaddressPublisher{}{Simon and Schuster}.
\PrintBackRefs{\CurrentBib}

\bibitem [\protect \citeauthoryear {%
Dusenbery%
}{%
Dusenbery%
}{%
{\protect \APACyear {1992}}%
}]{%
1008}
\APACinsertmetastar {%
1008}%
\begin{APACrefauthors}%
Dusenbery, D.B.%
\end{APACrefauthors}%
\unskip\
\newblock
\APACrefYear{1992}.
\newblock
\APACrefbtitle {Sensory ecology: how organisms acquire and respond to information} {Sensory ecology: how organisms acquire and respond to information}.
\newblock
\APACaddressPublisher{}{WH Freeman New York}.
\PrintBackRefs{\CurrentBib}

\bibitem [\protect \citeauthoryear {%
Dziri%
\ \protect \BOthers {.}}{%
Dziri%
\ \protect \BOthers {.}}{%
{\protect \APACyear {2023}}%
}]{%
785}
\APACinsertmetastar {%
785}%
\begin{APACrefauthors}%
Dziri, N.%
, Lu, X.%
, Sclar, M.%
, Li, X.L.%
, Jiang, L.%
, Lin, B.Y.%
\BDBL {}others%
\end{APACrefauthors}%
\unskip\
\newblock
\APACrefYearMonthDay{2023}{}{}.
\newblock
{\BBOQ}\APACrefatitle {Faith and fate: Limits of transformers on compositionality (2023)} {Faith and fate: Limits of transformers on compositionality (2023)}.{\BBCQ}
\newblock
\APACjournalVolNumPages{arXiv preprint arXiv:2305.18654}{}{}{}
\newblock

\newblock

\PrintBackRefs{\CurrentBib}

\bibitem [\protect \citeauthoryear {%
Frith%
}{%
Frith%
}{%
{\protect \APACyear {2007}}%
}]{%
989}
\APACinsertmetastar {%
989}%
\begin{APACrefauthors}%
Frith, C.%
\end{APACrefauthors}%
\unskip\
\newblock
\APACrefYear{2007}.
\newblock
\APACrefbtitle {Making up the mind: How the brain creates our mental world} {Making up the mind: How the brain creates our mental world}.
\newblock
\APACaddressPublisher{}{John Wiley \& Sons}.
\PrintBackRefs{\CurrentBib}

\bibitem [\protect \citeauthoryear {%
Gallagher%
\ \BBA {} Zahavi%
}{%
Gallagher%
\ \BBA {} Zahavi%
}{%
{\protect \APACyear {2020}}%
}]{%
983}
\APACinsertmetastar {%
983}%
\begin{APACrefauthors}%
Gallagher, S.%
\BCBT {}\ \BBA {} Zahavi, D.%
\end{APACrefauthors}%
\unskip\
\newblock
\APACrefYear{2020}.
\newblock
\APACrefbtitle {The phenomenological mind} {The phenomenological mind}.
\newblock
\APACaddressPublisher{}{Routledge}.
\PrintBackRefs{\CurrentBib}

\bibitem [\protect \citeauthoryear {%
Gibson%
}{%
Gibson%
}{%
{\protect \APACyear {2014}}%
}]{%
1012}
\APACinsertmetastar {%
1012}%
\begin{APACrefauthors}%
Gibson, J.J.%
\end{APACrefauthors}%
\unskip\
\newblock
\APACrefYear{2014}.
\newblock
\APACrefbtitle {The ecological approach to visual perception: classic edition} {The ecological approach to visual perception: classic edition}.
\newblock
\APACaddressPublisher{}{Psychology press}.
\PrintBackRefs{\CurrentBib}

\bibitem [\protect \citeauthoryear {%
Glasersfeld%
}{%
Glasersfeld%
}{%
{\protect \APACyear {1996}}%
}]{%
977}
\APACinsertmetastar {%
977}%
\begin{APACrefauthors}%
Glasersfeld, E.v.%
\end{APACrefauthors}%
\unskip\
\newblock
\APACrefYear{1996}.
\newblock
\APACrefbtitle {Radical constructivism: A way of knowing and learning} {Radical constructivism: A way of knowing and learning}.
\newblock
\APACaddressPublisher{}{Routledge}.
\PrintBackRefs{\CurrentBib}

\bibitem [\protect \citeauthoryear {%
Hatfield%
}{%
Hatfield%
}{%
{\protect \APACyear {1988}}%
}]{%
843}
\APACinsertmetastar {%
843}%
\begin{APACrefauthors}%
Hatfield, G.%
\end{APACrefauthors}%
\unskip\
\newblock
\APACrefYearMonthDay{1988}{}{}.
\newblock
{\BBOQ}\APACrefatitle {Representation and content in some (actual) theories of perception} {Representation and content in some (actual) theories of perception}.{\BBCQ}
\newblock
\APACjournalVolNumPages{Studies in History and Philosophy of Science Part A}{19}{2}{175--214}
\newblock

\newblock

\PrintBackRefs{\CurrentBib}

\bibitem [\protect \citeauthoryear {%
Haugeland%
}{%
Haugeland%
}{%
{\protect \APACyear {2000}}%
}]{%
979}
\APACinsertmetastar {%
979}%
\begin{APACrefauthors}%
Haugeland, J.%
\end{APACrefauthors}%
\unskip\
\newblock
\APACrefYear{2000}.
\newblock
\APACrefbtitle {Having thought: Essays in the metaphysics of mind} {Having thought: Essays in the metaphysics of mind}.
\newblock
\APACaddressPublisher{}{Harvard University Press}.
\PrintBackRefs{\CurrentBib}

\bibitem [\protect \citeauthoryear {%
Haugeland%
}{%
Haugeland%
}{%
{\protect \APACyear {2013}}%
}]{%
980}
\APACinsertmetastar {%
980}%
\begin{APACrefauthors}%
Haugeland, J.%
\end{APACrefauthors}%
\unskip\
\newblock
\APACrefYear{2013}.
\newblock
\APACrefbtitle {Dasein Disclosed: John Haugeland's Heidegger} {Dasein disclosed: John haugeland's heidegger}.
\newblock
\APACaddressPublisher{}{Harvard University Press}.
\PrintBackRefs{\CurrentBib}

\bibitem [\protect \citeauthoryear {%
Heidegger%
}{%
Heidegger%
}{%
{\protect \APACyear {1967}}%
}]{%
1001}
\APACinsertmetastar {%
1001}%
\begin{APACrefauthors}%
Heidegger, M.%
\end{APACrefauthors}%
\unskip\
\newblock
\APACrefYearMonthDay{1967}{}{}.
\newblock
\APACrefbtitle {Sein und Zeit.} {Sein und zeit.}
\newblock
\APACaddressPublisher{}{Max Niemeyer Verlag T{\"u}bingen}.
\PrintBackRefs{\CurrentBib}

\bibitem [\protect \citeauthoryear {%
Heidegger%
}{%
Heidegger%
}{%
{\protect \APACyear {1994}}%
}]{%
1015}
\APACinsertmetastar {%
1015}%
\begin{APACrefauthors}%
Heidegger, M.%
\end{APACrefauthors}%
\unskip\
\newblock
\APACrefYear{1994}.
\newblock
\APACrefbtitle {Ph{\"a}nomenologische Interpretationen zu Aristoteles} {Ph{\"a}nomenologische interpretationen zu aristoteles}.
\newblock
\APACaddressPublisher{}{Vittorio Klostermann}.
\PrintBackRefs{\CurrentBib}

\bibitem [\protect \citeauthoryear {%
Heidegger%
}{%
Heidegger%
}{%
{\protect \APACyear {1997}}%
}]{%
1018}
\APACinsertmetastar {%
1018}%
\begin{APACrefauthors}%
Heidegger, M.%
\end{APACrefauthors}%
\unskip\
\newblock
\APACrefYear{1997}.
\newblock
\APACrefbtitle {Der Satz vom Grund} {Der satz vom grund}.
\newblock
\APACaddressPublisher{}{Vittorio Klostermann}.
\PrintBackRefs{\CurrentBib}

\bibitem [\protect \citeauthoryear {%
Heidegger%
}{%
Heidegger%
}{%
{\protect \APACyear {1999}}%
}]{%
1019}
\APACinsertmetastar {%
1019}%
\begin{APACrefauthors}%
Heidegger, M.%
\end{APACrefauthors}%
\unskip\
\newblock
\APACrefYear{1999}.
\newblock
\APACrefbtitle {Metaphysik und Nihilismus} {Metaphysik und nihilismus}.
\newblock
\APACaddressPublisher{}{Vittorio Klostermann}.
\PrintBackRefs{\CurrentBib}

\bibitem [\protect \citeauthoryear {%
Heidegger%
}{%
Heidegger%
}{%
{\protect \APACyear {2000}}%
}]{%
1017}
\APACinsertmetastar {%
1017}%
\begin{APACrefauthors}%
Heidegger, M.%
\end{APACrefauthors}%
\unskip\
\newblock
\APACrefYear{2000}.
\newblock
\APACrefbtitle {{\"U}ber den Humanismus} {{\"U}ber den humanismus}.
\newblock
\APACaddressPublisher{}{Vittorio Klostermann}.
\PrintBackRefs{\CurrentBib}

\bibitem [\protect \citeauthoryear {%
Heidegger%
}{%
Heidegger%
}{%
{\protect \APACyear {2001}}%
}]{%
1020}
\APACinsertmetastar {%
1020}%
\begin{APACrefauthors}%
Heidegger, M.%
\end{APACrefauthors}%
\unskip\
\newblock
\APACrefYear{2001}.
\newblock
\APACrefbtitle {Sein und Wahrheit} {Sein und wahrheit}.
\newblock
\APACaddressPublisher{}{Vittorio Klostermann}.
\PrintBackRefs{\CurrentBib}

\bibitem [\protect \citeauthoryear {%
Heidegger%
}{%
Heidegger%
}{%
{\protect \APACyear {2012}}%
}]{%
1014}
\APACinsertmetastar {%
1014}%
\begin{APACrefauthors}%
Heidegger, M.%
\end{APACrefauthors}%
\unskip\
\newblock
\APACrefYear{2012}.
\newblock
\APACrefbtitle {Der Ursprung des Kunstwerkes} {Der ursprung des kunstwerkes}.
\newblock
\APACaddressPublisher{}{Vittorio Klostermann}.
\PrintBackRefs{\CurrentBib}

\bibitem [\protect \citeauthoryear {%
Hern{\'a}ndez-Orallo%
}{%
Hern{\'a}ndez-Orallo%
}{%
{\protect \APACyear {2017}}%
}]{%
799}
\APACinsertmetastar {%
799}%
\begin{APACrefauthors}%
Hern{\'a}ndez-Orallo, J.%
\end{APACrefauthors}%
\unskip\
\newblock
\APACrefYearMonthDay{2017}{}{}.
\newblock
{\BBOQ}\APACrefatitle {Evaluation in artificial intelligence: from task-oriented to ability-oriented measurement} {Evaluation in artificial intelligence: from task-oriented to ability-oriented measurement}.{\BBCQ}
\newblock
\APACjournalVolNumPages{Artificial Intelligence Review}{48}{}{397--447}
\newblock

\newblock

\PrintBackRefs{\CurrentBib}

\bibitem [\protect \citeauthoryear {%
Hoffstaetter%
, Bagriantsev%
\BCBL {}\ \BBA {} Gracheva%
}{%
Hoffstaetter%
\ \protect \BOthers {.}}{%
{\protect \APACyear {2018}}%
}]{%
807}
\APACinsertmetastar {%
807}%
\begin{APACrefauthors}%
Hoffstaetter, L.J.%
, Bagriantsev, S.N.%
\BCBL {} Gracheva, E.O.%
\end{APACrefauthors}%
\unskip\
\newblock
\APACrefYearMonthDay{2018}{}{}.
\newblock
{\BBOQ}\APACrefatitle {TRPs et al.: a molecular toolkit for thermosensory adaptations} {Trps et al.: a molecular toolkit for thermosensory adaptations}.{\BBCQ}
\newblock
\APACjournalVolNumPages{Pfl{\"u}gers Archiv-European Journal of Physiology}{470}{}{745--759}
\newblock

\newblock

\PrintBackRefs{\CurrentBib}

\bibitem [\protect \citeauthoryear {%
Horst%
}{%
Horst%
}{%
{\protect \APACyear {2005}}%
}]{%
738}
\APACinsertmetastar {%
738}%
\begin{APACrefauthors}%
Horst, S.%
\end{APACrefauthors}%
\unskip\
\newblock
\APACrefYearMonthDay{2005}{}{}.
\newblock
{\BBOQ}\APACrefatitle {Phenomenology and psychophysics} {Phenomenology and psychophysics}.{\BBCQ}
\newblock
\APACjournalVolNumPages{Phenomenology and the cognitive sciences}{4}{}{1--21}
\newblock

\newblock

\PrintBackRefs{\CurrentBib}

\bibitem [\protect \citeauthoryear {%
Hume%
}{%
Hume%
}{%
{\protect \APACyear {2016}}%
}]{%
1033}
\APACinsertmetastar {%
1033}%
\begin{APACrefauthors}%
Hume, D.%
\end{APACrefauthors}%
\unskip\
\newblock
\APACrefYearMonthDay{2016}{}{}.
\newblock
{\BBOQ}\APACrefatitle {An enquiry concerning human understanding} {An enquiry concerning human understanding}.{\BBCQ}
\newblock
 \APACrefbtitle {Seven masterpieces of philosophy} {Seven masterpieces of philosophy}\ (\BPGS\ 183--276).
\newblock
\APACaddressPublisher{}{Routledge}.
\PrintBackRefs{\CurrentBib}

\bibitem [\protect \citeauthoryear {%
Husserl%
}{%
Husserl%
}{%
{\protect \APACyear {1984}}%
}]{%
1016}
\APACinsertmetastar {%
1016}%
\begin{APACrefauthors}%
Husserl, E.%
\end{APACrefauthors}%
\unskip\
\newblock
\APACrefYear{1984}.
\newblock
\APACrefbtitle {Logische Untersuchungen. Zweiter Band, erster Teil.} {Logische untersuchungen. zweiter band, erster teil.}\ (U.~Panzer, \BED{}).
\newblock
\APACaddressPublisher{}{Springer}.
\PrintBackRefs{\CurrentBib}

\bibitem [\protect \citeauthoryear {%
Iriki%
, Tanaka%
\BCBL {}\ \BBA {} Iwamura%
}{%
Iriki%
\ \protect \BOthers {.}}{%
{\protect \APACyear {1996}}%
}]{%
842}
\APACinsertmetastar {%
842}%
\begin{APACrefauthors}%
Iriki, A.%
, Tanaka, M.%
\BCBL {} Iwamura, Y.%
\end{APACrefauthors}%
\unskip\
\newblock
\APACrefYearMonthDay{1996}{}{}.
\newblock
{\BBOQ}\APACrefatitle {Coding of modified body schema during tool use by macaque postcentral neurones.} {Coding of modified body schema during tool use by macaque postcentral neurones.}{\BBCQ}
\newblock
\APACjournalVolNumPages{Neuroreport}{7}{14}{2325--2330}
\newblock

\newblock

\PrintBackRefs{\CurrentBib}

\bibitem [\protect \citeauthoryear {%
Kant%
}{%
Kant%
}{%
{\protect \APACyear {1968}}%
}]{%
1029}
\APACinsertmetastar {%
1029}%
\begin{APACrefauthors}%
Kant, I.%
\end{APACrefauthors}%
\unskip\
\newblock
\APACrefYear{1968}.
\newblock
\APACrefbtitle {Kritik der reinen Vernunft} {Kritik der reinen vernunft}\ (\BVOL~3).
\newblock
\APACaddressPublisher{}{Suhrkamp}.
\PrintBackRefs{\CurrentBib}

\bibitem [\protect \citeauthoryear {%
Kelber%
, Vorobyev%
\BCBL {}\ \BBA {} Osorio%
}{%
Kelber%
\ \protect \BOthers {.}}{%
{\protect \APACyear {2003}}%
}]{%
805}
\APACinsertmetastar {%
805}%
\begin{APACrefauthors}%
Kelber, A.%
, Vorobyev, M.%
\BCBL {} Osorio, D.%
\end{APACrefauthors}%
\unskip\
\newblock
\APACrefYearMonthDay{2003}{}{}.
\newblock
{\BBOQ}\APACrefatitle {Animal colour vision--behavioural tests and physiological concepts} {Animal colour vision--behavioural tests and physiological concepts}.{\BBCQ}
\newblock
\APACjournalVolNumPages{Biological Reviews}{78}{1}{81--118}
\newblock

\newblock

\PrintBackRefs{\CurrentBib}

\bibitem [\protect \citeauthoryear {%
Keller%
, Zhuang%
, Chi%
, Vosshall%
\BCBL {}\ \BBA {} Matsunami%
}{%
Keller%
\ \protect \BOthers {.}}{%
{\protect \APACyear {2007}}%
}]{%
809}
\APACinsertmetastar {%
809}%
\begin{APACrefauthors}%
Keller, A.%
, Zhuang, H.%
, Chi, Q.%
, Vosshall, L.B.%
\BCBL {} Matsunami, H.%
\end{APACrefauthors}%
\unskip\
\newblock
\APACrefYearMonthDay{2007}{}{}.
\newblock
{\BBOQ}\APACrefatitle {Genetic variation in a human odorant receptor alters odour perception} {Genetic variation in a human odorant receptor alters odour perception}.{\BBCQ}
\newblock
\APACjournalVolNumPages{Nature}{449}{7161}{468--472}
\newblock

\newblock

\PrintBackRefs{\CurrentBib}

\bibitem [\protect \citeauthoryear {%
Khoram%
\ \protect \BOthers {.}}{%
Khoram%
\ \protect \BOthers {.}}{%
{\protect \APACyear {2019}}%
}]{%
835}
\APACinsertmetastar {%
835}%
\begin{APACrefauthors}%
Khoram, E.%
, Chen, A.%
, Liu, D.%
, Ying, L.%
, Wang, Q.%
, Yuan, M.%
\BCBL {} Yu, Z.%
\end{APACrefauthors}%
\unskip\
\newblock
\APACrefYearMonthDay{2019}{}{}.
\newblock
{\BBOQ}\APACrefatitle {Nanophotonic media for artificial neural inference} {Nanophotonic media for artificial neural inference}.{\BBCQ}
\newblock
\APACjournalVolNumPages{Photonics Research}{7}{8}{823--827}
\newblock

\newblock

\PrintBackRefs{\CurrentBib}

\bibitem [\protect \citeauthoryear {%
Lake%
, Ullman%
, Tenenbaum%
\BCBL {}\ \BBA {} Gershman%
}{%
Lake%
\ \protect \BOthers {.}}{%
{\protect \APACyear {2017}}%
}]{%
282}
\APACinsertmetastar {%
282}%
\begin{APACrefauthors}%
Lake, B.M.%
, Ullman, T.D.%
, Tenenbaum, J.B.%
\BCBL {} Gershman, S.J.%
\end{APACrefauthors}%
\unskip\
\newblock
\APACrefYearMonthDay{2017}{}{}.
\newblock
{\BBOQ}\APACrefatitle {Building machines that learn and think like people} {Building machines that learn and think like people}.{\BBCQ}
\newblock
\APACjournalVolNumPages{Behavioral and brain sciences}{40}{}{e253}
\newblock

\newblock

\PrintBackRefs{\CurrentBib}

\bibitem [\protect \citeauthoryear {%
Legg%
, Hutter%
\BCBL {}\ \protect \BOthers {.}}{%
Legg%
\ \protect \BOthers {.}}{%
{\protect \APACyear {2007}}%
}]{%
798}
\APACinsertmetastar {%
798}%
\begin{APACrefauthors}%
Legg, S.%
, Hutter, M.%
\BCBL {}\ \BOthersPeriod {.}\end{APACrefauthors}%
\unskip\
\newblock
\APACrefYearMonthDay{2007}{}{}.
\newblock
{\BBOQ}\APACrefatitle {A collection of definitions of intelligence} {A collection of definitions of intelligence}.{\BBCQ}
\newblock
\APACjournalVolNumPages{Frontiers in Artificial Intelligence and applications}{157}{}{17}
\newblock

\newblock

\PrintBackRefs{\CurrentBib}

\bibitem [\protect \citeauthoryear {%
Lhermitte%
}{%
Lhermitte%
}{%
{\protect \APACyear {1986}}%
}]{%
836}
\APACinsertmetastar {%
836}%
\begin{APACrefauthors}%
Lhermitte, F.%
\end{APACrefauthors}%
\unskip\
\newblock
\APACrefYearMonthDay{1986}{}{}.
\newblock
{\BBOQ}\APACrefatitle {Human autonomy and the frontal lobes. Part II: patient behavior in complex and social situations: the “environmental dependency syndrome”} {Human autonomy and the frontal lobes. part ii: patient behavior in complex and social situations: the “environmental dependency syndrome”}.{\BBCQ}
\newblock
\APACjournalVolNumPages{Annals of Neurology: Official Journal of the American Neurological Association and the Child Neurology Society}{19}{4}{335--343}
\newblock

\newblock

\PrintBackRefs{\CurrentBib}

\bibitem [\protect \citeauthoryear {%
Locke%
}{%
Locke%
}{%
{\protect \APACyear {1847}}%
}]{%
1028}
\APACinsertmetastar {%
1028}%
\begin{APACrefauthors}%
Locke, J.%
\end{APACrefauthors}%
\unskip\
\newblock
\APACrefYear{1847}.
\newblock
\APACrefbtitle {An essay concerning human understanding} {An essay concerning human understanding}.
\newblock
\APACaddressPublisher{}{Kay \& Troutman}.
\PrintBackRefs{\CurrentBib}

\bibitem [\protect \citeauthoryear {%
Ma%
, Tsao%
\BCBL {}\ \BBA {} Shum%
}{%
Ma%
\ \protect \BOthers {.}}{%
{\protect \APACyear {2022}}%
}]{%
766}
\APACinsertmetastar {%
766}%
\begin{APACrefauthors}%
Ma, Y.%
, Tsao, D.%
\BCBL {} Shum, H\BHBI Y.%
\end{APACrefauthors}%
\unskip\
\newblock
\APACrefYearMonthDay{2022}{}{}.
\newblock
{\BBOQ}\APACrefatitle {On the principles of parsimony and self-consistency for the emergence of intelligence} {On the principles of parsimony and self-consistency for the emergence of intelligence}.{\BBCQ}
\newblock
\APACjournalVolNumPages{Frontiers of Information Technology \& Electronic Engineering}{23}{9}{1298--1323}
\newblock

\newblock

\PrintBackRefs{\CurrentBib}

\bibitem [\protect \citeauthoryear {%
MacIver%
, Schmitz%
, Mugan%
, Murphey%
\BCBL {}\ \BBA {} Mobley%
}{%
MacIver%
\ \protect \BOthers {.}}{%
{\protect \APACyear {2017}}%
}]{%
838}
\APACinsertmetastar {%
838}%
\begin{APACrefauthors}%
MacIver, M.A.%
, Schmitz, L.%
, Mugan, U.%
, Murphey, T.D.%
\BCBL {} Mobley, C.D.%
\end{APACrefauthors}%
\unskip\
\newblock
\APACrefYearMonthDay{2017}{}{}.
\newblock
{\BBOQ}\APACrefatitle {Massive increase in visual range preceded the origin of terrestrial vertebrates} {Massive increase in visual range preceded the origin of terrestrial vertebrates}.{\BBCQ}
\newblock
\APACjournalVolNumPages{Proceedings of the National Academy of Sciences}{114}{12}{E2375--E2384}
\newblock

\newblock

\PrintBackRefs{\CurrentBib}

\bibitem [\protect \citeauthoryear {%
McCoy%
, Yao%
, Friedman%
, Hardy%
\BCBL {}\ \BBA {} Griffiths%
}{%
McCoy%
\ \protect \BOthers {.}}{%
{\protect \APACyear {2023}}%
}]{%
784}
\APACinsertmetastar {%
784}%
\begin{APACrefauthors}%
McCoy, R.T.%
, Yao, S.%
, Friedman, D.%
, Hardy, M.%
\BCBL {} Griffiths, T.L.%
\end{APACrefauthors}%
\unskip\
\newblock
\APACrefYearMonthDay{2023}{}{}.
\newblock
{\BBOQ}\APACrefatitle {Embers of autoregression: Understanding large language models through the problem they are trained to solve} {Embers of autoregression: Understanding large language models through the problem they are trained to solve}.{\BBCQ}
\newblock
\APACjournalVolNumPages{arXiv preprint arXiv:2309.13638}{}{}{}
\newblock

\newblock

\PrintBackRefs{\CurrentBib}

\bibitem [\protect \citeauthoryear {%
Merleau-Ponty%
}{%
Merleau-Ponty%
}{%
{\protect \APACyear {2012}}%
}]{%
1013}
\APACinsertmetastar {%
1013}%
\begin{APACrefauthors}%
Merleau-Ponty, M.%
\end{APACrefauthors}%
\unskip\
\newblock
\APACrefYear{2012}.
\newblock
\APACrefbtitle {Phenomenology of Perception} {Phenomenology of perception}.
\newblock
\APACaddressPublisher{}{Routledge}.
\PrintBackRefs{\CurrentBib}

\bibitem [\protect \citeauthoryear {%
Mitchell%
}{%
Mitchell%
}{%
{\protect \APACyear {2021}}%
}]{%
796}
\APACinsertmetastar {%
796}%
\begin{APACrefauthors}%
Mitchell, M.%
\end{APACrefauthors}%
\unskip\
\newblock
\APACrefYearMonthDay{2021}{}{}.
\newblock
{\BBOQ}\APACrefatitle {Why AI is harder than we think} {Why ai is harder than we think}.{\BBCQ}
\newblock
\APACjournalVolNumPages{arXiv preprint arXiv:2104.12871}{}{}{}
\newblock

\newblock

\PrintBackRefs{\CurrentBib}

\bibitem [\protect \citeauthoryear {%
Neurath%
}{%
Neurath%
}{%
{\protect \APACyear {1932}}%
}]{%
830}
\APACinsertmetastar {%
830}%
\begin{APACrefauthors}%
Neurath, O.%
\end{APACrefauthors}%
\unskip\
\newblock
\APACrefYearMonthDay{1932}{}{}.
\newblock
{\BBOQ}\APACrefatitle {Protokolls{\"a}tze} {Protokolls{\"a}tze}.{\BBCQ}
\newblock
\APACjournalVolNumPages{Erkenntnis}{3}{}{204--214}
\newblock

\newblock

\PrintBackRefs{\CurrentBib}

\bibitem [\protect \citeauthoryear {%
Newman%
, Choi%
, Wynn%
\BCBL {}\ \BBA {} Scholl%
}{%
Newman%
\ \protect \BOthers {.}}{%
{\protect \APACyear {2008}}%
}]{%
291}
\APACinsertmetastar {%
291}%
\begin{APACrefauthors}%
Newman, G.E.%
, Choi, H.%
, Wynn, K.%
\BCBL {} Scholl, B.J.%
\end{APACrefauthors}%
\unskip\
\newblock
\APACrefYearMonthDay{2008}{}{}.
\newblock
{\BBOQ}\APACrefatitle {The origins of causal perception: Evidence from postdictive processing in infancy} {The origins of causal perception: Evidence from postdictive processing in infancy}.{\BBCQ}
\newblock
\APACjournalVolNumPages{Cognitive psychology}{57}{3}{262--291}
\newblock

\newblock

\PrintBackRefs{\CurrentBib}

\bibitem [\protect \citeauthoryear {%
Nezhurina%
, Cipolina-Kun%
, Cherti%
\BCBL {}\ \BBA {} Jitsev%
}{%
Nezhurina%
\ \protect \BOthers {.}}{%
{\protect \APACyear {2024}}%
}]{%
783}
\APACinsertmetastar {%
783}%
\begin{APACrefauthors}%
Nezhurina, M.%
, Cipolina-Kun, L.%
, Cherti, M.%
\BCBL {} Jitsev, J.%
\end{APACrefauthors}%
\unskip\
\newblock
\APACrefYearMonthDay{2024}{}{}.
\newblock
{\BBOQ}\APACrefatitle {Alice in Wonderland: Simple Tasks Showing Complete Reasoning Breakdown in State-Of-the-Art Large Language Models} {Alice in wonderland: Simple tasks showing complete reasoning breakdown in state-of-the-art large language models}.{\BBCQ}
\newblock
\APACjournalVolNumPages{arXiv preprint arXiv:2406.02061}{}{}{}
\newblock

\newblock

\PrintBackRefs{\CurrentBib}

\bibitem [\protect \citeauthoryear {%
Nietzsche%
}{%
Nietzsche%
}{%
{\protect \APACyear {1982}}%
}]{%
831}
\APACinsertmetastar {%
831}%
\begin{APACrefauthors}%
Nietzsche, F.%
\end{APACrefauthors}%
\unskip\
\newblock
\APACrefYear{1982}.
\newblock
\APACrefbtitle {{\"U}ber Wahrheit und L{\"u}ge im au{\ss}ermoralischen Sinne} {{\"U}ber wahrheit und l{\"u}ge im au{\ss}ermoralischen sinne}.
\newblock
\APACaddressPublisher{}{Carl Hanser Verlag, M{\"u}nchen}.
\PrintBackRefs{\CurrentBib}

\bibitem [\protect \citeauthoryear {%
Nilsson%
}{%
Nilsson%
}{%
{\protect \APACyear {2009}}%
}]{%
804}
\APACinsertmetastar {%
804}%
\begin{APACrefauthors}%
Nilsson, D\BHBI E.%
\end{APACrefauthors}%
\unskip\
\newblock
\APACrefYearMonthDay{2009}{}{}.
\newblock
{\BBOQ}\APACrefatitle {The evolution of eyes and visually guided behaviour} {The evolution of eyes and visually guided behaviour}.{\BBCQ}
\newblock
\APACjournalVolNumPages{Philosophical Transactions of the Royal Society B: Biological Sciences}{364}{1531}{2833--2847}
\newblock

\newblock

\PrintBackRefs{\CurrentBib}

\bibitem [\protect \citeauthoryear {%
Okasha%
}{%
Okasha%
}{%
{\protect \APACyear {2001}}%
}]{%
661}
\APACinsertmetastar {%
661}%
\begin{APACrefauthors}%
Okasha, S.%
\end{APACrefauthors}%
\unskip\
\newblock
\APACrefYearMonthDay{2001}{}{}.
\newblock
{\BBOQ}\APACrefatitle {What did hume really show about induction?} {What did hume really show about induction?}{\BBCQ}
\newblock
\APACjournalVolNumPages{The Philosophical Quarterly}{51}{204}{307--327}
\newblock

\newblock

\PrintBackRefs{\CurrentBib}

\bibitem [\protect \citeauthoryear {%
Peirce%
}{%
Peirce%
}{%
{\protect \APACyear {1958}}%
}]{%
1022}
\APACinsertmetastar {%
1022}%
\begin{APACrefauthors}%
Peirce, C.S.%
\end{APACrefauthors}%
\unskip\
\newblock
\APACrefYear{1958}.
\newblock
\APACrefbtitle {Collected papers of charles sanders peirce. 8 volumes. Edited by C. Hartshorne, \& P. Weiss (Eds.) (Vol. 1–6) and A.W. Burks (Ed.) (Vol. 7-8)} {Collected papers of charles sanders peirce. 8 volumes. edited by c. hartshorne, \& p. weiss (eds.) (vol. 1–6) and a.w. burks (ed.) (vol. 7-8)}.
\newblock
\APACaddressPublisher{}{Cambridge MA: Belknap Press}.
\PrintBackRefs{\CurrentBib}

\bibitem [\protect \citeauthoryear {%
Peirce%
}{%
Peirce%
}{%
{\protect \APACyear {1998}}%
}]{%
1023}
\APACinsertmetastar {%
1023}%
\begin{APACrefauthors}%
Peirce, C.S.%
\end{APACrefauthors}%
\unskip\
\newblock
\APACrefYear{1998}.
\newblock
\APACrefbtitle {The Essential Peirce, Vol. 2: Selected Philosophical Writings (1893-1913)} {The essential peirce, vol. 2: Selected philosophical writings (1893-1913)}.
\newblock
\APACaddressPublisher{}{Indiana University Press}.
\PrintBackRefs{\CurrentBib}

\bibitem [\protect \citeauthoryear {%
Pfeifer%
\ \BBA {} Bongard%
}{%
Pfeifer%
\ \BBA {} Bongard%
}{%
{\protect \APACyear {2006}}%
}]{%
976}
\APACinsertmetastar {%
976}%
\begin{APACrefauthors}%
Pfeifer, R.%
\BCBT {}\ \BBA {} Bongard, J.%
\end{APACrefauthors}%
\unskip\
\newblock
\APACrefYear{2006}.
\newblock
\APACrefbtitle {How the body shapes the way we think: a new view of intelligence} {How the body shapes the way we think: a new view of intelligence}.
\newblock
\APACaddressPublisher{}{MIT press}.
\PrintBackRefs{\CurrentBib}

\bibitem [\protect \citeauthoryear {%
Pfister%
}{%
Pfister%
}{%
{\protect \APACyear {2022}}%
}]{%
711}
\APACinsertmetastar {%
711}%
\begin{APACrefauthors}%
Pfister, R.%
\end{APACrefauthors}%
\unskip\
\newblock
\APACrefYearMonthDay{2022}{}{}.
\newblock
{\BBOQ}\APACrefatitle {Towards a theory of abduction based on conditionals} {Towards a theory of abduction based on conditionals}.{\BBCQ}
\newblock
\APACjournalVolNumPages{Synthese}{200}{3}{206}
\newblock

\newblock

\PrintBackRefs{\CurrentBib}

\bibitem [\protect \citeauthoryear {%
Pfister%
}{%
Pfister%
}{%
{\protect \APACyear {2025}}%
}]{%
829}
\APACinsertmetastar {%
829}%
\begin{APACrefauthors}%
Pfister, R.%
\end{APACrefauthors}%
\unskip\
\newblock
\APACrefYearMonthDay{2025}{}{}.
\newblock
{\BBOQ}\APACrefatitle {Beyond Theory: A Philosophical Framework for Decision-Making in Management} {Beyond theory: A philosophical framework for decision-making in management}.{\BBCQ}
\newblock
 C.H.~Hoffmann\ (\BED), \APACrefbtitle {Artificial Intelligence, Entrepreneurship and Risk: Reflections and Positions at the Crossroads between Philosophy and Management.} {Artificial intelligence, entrepreneurship and risk: Reflections and positions at the crossroads between philosophy and management.}
\newblock
\APACaddressPublisher{}{Springer VS Wiesbaden}.
\newblock
\APACrefnote{Due April 8, 2025}
\PrintBackRefs{\CurrentBib}

\bibitem [\protect \citeauthoryear {%
Pfister%
\ \BBA {} Jud%
}{%
Pfister%
\ \BBA {} Jud%
}{%
{\protect \APACyear {2025}}%
}]{%
841}
\APACinsertmetastar {%
841}%
\begin{APACrefauthors}%
Pfister, R.%
\BCBT {}\ \BBA {} Jud, H.%
\end{APACrefauthors}%
\unskip\
\newblock
\APACrefYearMonthDay{2025}{}{}.
\newblock
{\BBOQ}\APACrefatitle {Understanding and Benchmarking Artificial Intelligence: OpenAI's o3 Is Not AGI} {Understanding and benchmarking artificial intelligence: Openai's o3 is not agi}.{\BBCQ}
\newblock
\APACjournalVolNumPages{arXiv preprint arXiv:2501.07458}{}{}{}
\newblock

\newblock

\PrintBackRefs{\CurrentBib}

\bibitem [\protect \citeauthoryear {%
Porter%
\ \protect \BOthers {.}}{%
Porter%
\ \protect \BOthers {.}}{%
{\protect \APACyear {2012}}%
}]{%
810}
\APACinsertmetastar {%
810}%
\begin{APACrefauthors}%
Porter, M.L.%
, Blasic, J.R.%
, Bok, M.J.%
, Cameron, E.G.%
, Pringle, T.%
, Cronin, T.W.%
\BCBL {} Robinson, P.R.%
\end{APACrefauthors}%
\unskip\
\newblock
\APACrefYearMonthDay{2012}{}{}.
\newblock
{\BBOQ}\APACrefatitle {Shedding new light on opsin evolution} {Shedding new light on opsin evolution}.{\BBCQ}
\newblock
\APACjournalVolNumPages{Proceedings of the Royal Society B: Biological Sciences}{279}{1726}{3--14}
\newblock

\newblock

\PrintBackRefs{\CurrentBib}

\bibitem [\protect \citeauthoryear {%
Preston%
}{%
Preston%
}{%
{\protect \APACyear {1993}}%
}]{%
768}
\APACinsertmetastar {%
768}%
\begin{APACrefauthors}%
Preston, B.%
\end{APACrefauthors}%
\unskip\
\newblock
\APACrefYearMonthDay{1993}{}{}.
\newblock
{\BBOQ}\APACrefatitle {Heidegger and artificial intelligence} {Heidegger and artificial intelligence}.{\BBCQ}
\newblock
\APACjournalVolNumPages{Philosophy and Phenomenological Research}{53}{1}{43--69}
\newblock

\newblock

\PrintBackRefs{\CurrentBib}

\bibitem [\protect \citeauthoryear {%
Putnam%
\ \protect \BOthers {.}}{%
Putnam%
\ \protect \BOthers {.}}{%
{\protect \APACyear {1981}}%
}]{%
1030}
\APACinsertmetastar {%
1030}%
\begin{APACrefauthors}%
Putnam, H.%
\BCBT {}\ \BOthersPeriod {.}
\end{APACrefauthors}%
\unskip\
\newblock
\APACrefYear{1981}.
\newblock
\APACrefbtitle {Reason, truth and history} {Reason, truth and history}\ (\BVOL~3).
\newblock
\APACaddressPublisher{}{Cambridge University Press Cambridge}.
\PrintBackRefs{\CurrentBib}

\bibitem [\protect \citeauthoryear {%
Rosenthal%
}{%
Rosenthal%
}{%
{\protect \APACyear {2004}}%
}]{%
834}
\APACinsertmetastar {%
834}%
\begin{APACrefauthors}%
Rosenthal, S.%
\end{APACrefauthors}%
\unskip\
\newblock
\APACrefYearMonthDay{2004}{}{}.
\newblock
{\BBOQ}\APACrefatitle {Peirce’s pragmatic account of perception: Issues and implications} {Peirce’s pragmatic account of perception: Issues and implications}.{\BBCQ}
\newblock
\APACjournalVolNumPages{The Cambridge Companion to Peirce}{}{}{193--213}
\newblock

\newblock

\PrintBackRefs{\CurrentBib}

\bibitem [\protect \citeauthoryear {%
Russell%
\ \BBA {} Norvig%
}{%
Russell%
\ \BBA {} Norvig%
}{%
{\protect \APACyear {2022}}%
}]{%
1007}
\APACinsertmetastar {%
1007}%
\begin{APACrefauthors}%
Russell, S.J.%
\BCBT {}\ \BBA {} Norvig, P.%
\end{APACrefauthors}%
\unskip\
\newblock
\APACrefYear{2022}.
\newblock
\APACrefbtitle {Artificial intelligence: A modern approach, global edition} {Artificial intelligence: A modern approach, global edition}.
\newblock
\APACaddressPublisher{}{Harlow: Pearson}.
\PrintBackRefs{\CurrentBib}

\bibitem [\protect \citeauthoryear {%
Safranski%
}{%
Safranski%
}{%
{\protect \APACyear {2014}}%
}]{%
999}
\APACinsertmetastar {%
999}%
\begin{APACrefauthors}%
Safranski, R.%
\end{APACrefauthors}%
\unskip\
\newblock
\APACrefYear{2014}.
\newblock
\APACrefbtitle {Ein Meister aus Deutschland: Heidegger und seine Zeit} {Ein meister aus deutschland: Heidegger und seine zeit}.
\newblock
\APACaddressPublisher{}{Carl Hanser Verlag}.
\PrintBackRefs{\CurrentBib}

\bibitem [\protect \citeauthoryear {%
Schaffer%
}{%
Schaffer%
}{%
{\protect \APACyear {2005}}%
}]{%
800}
\APACinsertmetastar {%
800}%
\begin{APACrefauthors}%
Schaffer, J.%
\end{APACrefauthors}%
\unskip\
\newblock
\APACrefYearMonthDay{2005}{}{}.
\newblock
{\BBOQ}\APACrefatitle {Contrastive knowledge} {Contrastive knowledge}.{\BBCQ}
\newblock
\APACjournalVolNumPages{Oxford studies in epistemology}{1}{}{235--271}
\newblock

\newblock

\PrintBackRefs{\CurrentBib}

\bibitem [\protect \citeauthoryear {%
Schneider%
\ \BBA {} McGrew%
}{%
Schneider%
\ \BBA {} McGrew%
}{%
{\protect \APACyear {2018}}%
}]{%
797}
\APACinsertmetastar {%
797}%
\begin{APACrefauthors}%
Schneider, W.J.%
\BCBT {}\ \BBA {} McGrew, K.S.%
\end{APACrefauthors}%
\unskip\
\newblock
\APACrefYearMonthDay{2018}{}{}.
\newblock
{\BBOQ}\APACrefatitle {The Cattell-Horn-Carroll theory of cognitive abilities} {The cattell-horn-carroll theory of cognitive abilities}.{\BBCQ}
\newblock
\APACjournalVolNumPages{Contemporary intellectual assessment: Theories, tests, and issues}{}{}{73--163}
\newblock

\newblock

\PrintBackRefs{\CurrentBib}

\bibitem [\protect \citeauthoryear {%
Searle%
}{%
Searle%
}{%
{\protect \APACyear {1980}}%
}]{%
814}
\APACinsertmetastar {%
814}%
\begin{APACrefauthors}%
Searle, J.R.%
\end{APACrefauthors}%
\unskip\
\newblock
\APACrefYearMonthDay{1980}{}{}.
\newblock
{\BBOQ}\APACrefatitle {Minds, brains, and programs} {Minds, brains, and programs}.{\BBCQ}
\newblock
\APACjournalVolNumPages{Behavioral and brain sciences}{3}{3}{417--424}
\newblock

\newblock

\PrintBackRefs{\CurrentBib}

\bibitem [\protect \citeauthoryear {%
Searle%
}{%
Searle%
}{%
{\protect \APACyear {1992}}%
}]{%
1027}
\APACinsertmetastar {%
1027}%
\begin{APACrefauthors}%
Searle, J.R.%
\end{APACrefauthors}%
\unskip\
\newblock
\APACrefYear{1992}.
\newblock
\APACrefbtitle {The rediscovery of the mind} {The rediscovery of the mind}.
\newblock
\APACaddressPublisher{}{MIT press}.
\PrintBackRefs{\CurrentBib}

\bibitem [\protect \citeauthoryear {%
Seff%
\ \protect \BOthers {.}}{%
Seff%
\ \protect \BOthers {.}}{%
{\protect \APACyear {2023}}%
}]{%
790}
\APACinsertmetastar {%
790}%
\begin{APACrefauthors}%
Seff, A.%
, Cera, B.%
, Chen, D.%
, Ng, M.%
, Zhou, A.%
, Nayakanti, N.%
\BDBL {}Sapp, B.%
\end{APACrefauthors}%
\unskip\
\newblock
\APACrefYearMonthDay{2023}{}{}.
\newblock
{\BBOQ}\APACrefatitle {Motionlm: Multi-agent motion forecasting as language modeling} {Motionlm: Multi-agent motion forecasting as language modeling}.{\BBCQ}
\newblock
 \APACrefbtitle {Proceedings of the IEEE/CVF International Conference on Computer Vision} {Proceedings of the ieee/cvf international conference on computer vision}\ (\BPGS\ 8579--8590).
\PrintBackRefs{\CurrentBib}

\bibitem [\protect \citeauthoryear {%
Shanahan%
}{%
Shanahan%
}{%
{\protect \APACyear {2005}}%
}]{%
437}
\APACinsertmetastar {%
437}%
\begin{APACrefauthors}%
Shanahan, M.%
\end{APACrefauthors}%
\unskip\
\newblock
\APACrefYearMonthDay{2005}{}{}.
\newblock
{\BBOQ}\APACrefatitle {Perception as abduction: Turning sensor data into meaningful representation} {Perception as abduction: Turning sensor data into meaningful representation}.{\BBCQ}
\newblock
\APACjournalVolNumPages{Cognitive science}{29}{1}{103--134}
\newblock

\newblock

\PrintBackRefs{\CurrentBib}

\bibitem [\protect \citeauthoryear {%
Shanahan%
\ \BBA {} Mitchell%
}{%
Shanahan%
\ \BBA {} Mitchell%
}{%
{\protect \APACyear {2022}}%
}]{%
795}
\APACinsertmetastar {%
795}%
\begin{APACrefauthors}%
Shanahan, M.%
\BCBT {}\ \BBA {} Mitchell, M.%
\end{APACrefauthors}%
\unskip\
\newblock
\APACrefYearMonthDay{2022}{}{}.
\newblock
{\BBOQ}\APACrefatitle {Abstraction for deep reinforcement learning} {Abstraction for deep reinforcement learning}.{\BBCQ}
\newblock
\APACjournalVolNumPages{arXiv preprint arXiv:2202.05839}{}{}{}
\newblock

\newblock

\PrintBackRefs{\CurrentBib}

\bibitem [\protect \citeauthoryear {%
Shapiro%
}{%
Shapiro%
}{%
{\protect \APACyear {2019}}%
}]{%
998}
\APACinsertmetastar {%
998}%
\begin{APACrefauthors}%
Shapiro, L.%
\end{APACrefauthors}%
\unskip\
\newblock
\APACrefYear{2019}.
\newblock
\APACrefbtitle {Embodied cognition} {Embodied cognition}.
\newblock
\APACaddressPublisher{}{Routledge}.
\PrintBackRefs{\CurrentBib}

\bibitem [\protect \citeauthoryear {%
Skinner%
}{%
Skinner%
}{%
{\protect \APACyear {1948}}%
}]{%
844}
\APACinsertmetastar {%
844}%
\begin{APACrefauthors}%
Skinner, B.F.%
\end{APACrefauthors}%
\unskip\
\newblock
\APACrefYearMonthDay{1948}{}{}.
\newblock
{\BBOQ}\APACrefatitle {'Superstition'in the pigeon.} {'superstition'in the pigeon.}{\BBCQ}
\newblock
\APACjournalVolNumPages{Journal of experimental psychology}{38}{2}{168}
\newblock

\newblock

\PrintBackRefs{\CurrentBib}

\bibitem [\protect \citeauthoryear {%
Smith%
}{%
Smith%
}{%
{\protect \APACyear {2018}}%
}]{%
741}
\APACinsertmetastar {%
741}%
\begin{APACrefauthors}%
Smith, D.W.%
\end{APACrefauthors}%
\unskip\
\newblock
\APACrefYearMonthDay{2018}{}{}.
\newblock
{\BBOQ}\APACrefatitle {{Phenomenology}} {{Phenomenology}}.{\BBCQ}
\newblock
 E.N.~Zalta\ (\BED), \APACrefbtitle {The {Stanford} Encyclopedia of Philosophy} {The {Stanford} encyclopedia of philosophy}\ (\PrintOrdinal{{S}ummer 2018}\ \BEd).
\newblock
\APACaddressPublisher{}{Metaphysics Research Lab, Stanford University}.
\PrintBackRefs{\CurrentBib}

\bibitem [\protect \citeauthoryear {%
Spelke%
}{%
Spelke%
}{%
{\protect \APACyear {2022}}%
}]{%
975}
\APACinsertmetastar {%
975}%
\begin{APACrefauthors}%
Spelke, E.%
\end{APACrefauthors}%
\unskip\
\newblock
\APACrefYear{2022}.
\newblock
\APACrefbtitle {What babies know: Core knowledge and composition volume 1} {What babies know: Core knowledge and composition volume 1}\ (\BVOL~1).
\newblock
\APACaddressPublisher{}{Oxford University Press}.
\PrintBackRefs{\CurrentBib}

\bibitem [\protect \citeauthoryear {%
Suk%
, Lee%
, Kim%
\BCBL {}\ \BBA {} Kim%
}{%
Suk%
\ \protect \BOthers {.}}{%
{\protect \APACyear {2024}}%
}]{%
791}
\APACinsertmetastar {%
791}%
\begin{APACrefauthors}%
Suk, H.%
, Lee, Y.%
, Kim, T.%
\BCBL {} Kim, S.%
\end{APACrefauthors}%
\unskip\
\newblock
\APACrefYearMonthDay{2024}{}{}.
\newblock
{\BBOQ}\APACrefatitle {Addressing uncertainty challenges for autonomous driving in real-world environments} {Addressing uncertainty challenges for autonomous driving in real-world environments}.{\BBCQ}
\newblock
 \APACrefbtitle {Advances in Computers} {Advances in computers}\ (\BVOL~134, \BPGS\ 317--361).
\newblock
\APACaddressPublisher{}{Elsevier}.
\PrintBackRefs{\CurrentBib}

\bibitem [\protect \citeauthoryear {%
Tenenbaum%
, Kemp%
, Griffiths%
\BCBL {}\ \BBA {} Goodman%
}{%
Tenenbaum%
\ \protect \BOthers {.}}{%
{\protect \APACyear {2011}}%
}]{%
206}
\APACinsertmetastar {%
206}%
\begin{APACrefauthors}%
Tenenbaum, J.B.%
, Kemp, C.%
, Griffiths, T.L.%
\BCBL {} Goodman, N.D.%
\end{APACrefauthors}%
\unskip\
\newblock
\APACrefYearMonthDay{2011}{}{}.
\newblock
{\BBOQ}\APACrefatitle {How to grow a mind: Statistics, structure, and abstraction} {How to grow a mind: Statistics, structure, and abstraction}.{\BBCQ}
\newblock
\APACjournalVolNumPages{science}{331}{6022}{1279--1285}
\newblock

\newblock

\PrintBackRefs{\CurrentBib}

\bibitem [\protect \citeauthoryear {%
Thagard%
}{%
Thagard%
}{%
{\protect \APACyear {2012}}%
}]{%
754}
\APACinsertmetastar {%
754}%
\begin{APACrefauthors}%
Thagard, P.%
\end{APACrefauthors}%
\unskip\
\newblock
\APACrefYearMonthDay{2012}{}{}.
\newblock
{\BBOQ}\APACrefatitle {Creative combination of representations: Scientific discovery and technological invention} {Creative combination of representations: Scientific discovery and technological invention}.{\BBCQ}
\newblock
\APACjournalVolNumPages{Psychology of science: Implicit and explicit processes}{}{}{389--405}
\newblock

\newblock

\PrintBackRefs{\CurrentBib}

\bibitem [\protect \citeauthoryear {%
Touvron%
\ \protect \BOthers {.}}{%
Touvron%
\ \protect \BOthers {.}}{%
{\protect \APACyear {2023}}%
}]{%
788}
\APACinsertmetastar {%
788}%
\begin{APACrefauthors}%
Touvron, H.%
, Lavril, T.%
, Izacard, G.%
, Martinet, X.%
, Lachaux, M\BHBI A.%
, Lacroix, T.%
\BDBL {}others%
\end{APACrefauthors}%
\unskip\
\newblock
\APACrefYearMonthDay{2023}{}{}.
\newblock
{\BBOQ}\APACrefatitle {Llama: Open and efficient foundation language models} {Llama: Open and efficient foundation language models}.{\BBCQ}
\newblock
\APACjournalVolNumPages{arXiv preprint arXiv:2302.13971}{}{}{}
\newblock

\newblock

\PrintBackRefs{\CurrentBib}

\bibitem [\protect \citeauthoryear {%
Tye%
}{%
Tye%
}{%
{\protect \APACyear {2021}}%
}]{%
845}
\APACinsertmetastar {%
845}%
\begin{APACrefauthors}%
Tye, M.%
\end{APACrefauthors}%
\unskip\
\newblock
\APACrefYearMonthDay{2021}{}{}.
\newblock
{\BBOQ}\APACrefatitle {{Qualia}} {{Qualia}}.{\BBCQ}
\newblock
 E.N.~Zalta\ (\BED), \APACrefbtitle {The {Stanford} Encyclopedia of Philosophy} {The {Stanford} encyclopedia of philosophy}\ (\PrintOrdinal{{F}all 2021}\ \BEd).
\newblock
\APACaddressPublisher{}{Metaphysics Research Lab, Stanford University}.
\PrintBackRefs{\CurrentBib}

\bibitem [\protect \citeauthoryear {%
Valentin{\v{c}}i{\v{c}}%
, Wegert%
\BCBL {}\ \BBA {} Caprio%
}{%
Valentin{\v{c}}i{\v{c}}%
\ \protect \BOthers {.}}{%
{\protect \APACyear {1994}}%
}]{%
808}
\APACinsertmetastar {%
808}%
\begin{APACrefauthors}%
Valentin{\v{c}}i{\v{c}}, T.%
, Wegert, S.%
\BCBL {} Caprio, J.%
\end{APACrefauthors}%
\unskip\
\newblock
\APACrefYearMonthDay{1994}{}{}.
\newblock
{\BBOQ}\APACrefatitle {Learned olfactory discrimination versus innate taste responses to amino acids in channel catfish (Ictalurus punctatus)} {Learned olfactory discrimination versus innate taste responses to amino acids in channel catfish (ictalurus punctatus)}.{\BBCQ}
\newblock
\APACjournalVolNumPages{Physiology \& behavior}{55}{5}{865--873}
\newblock

\newblock

\PrintBackRefs{\CurrentBib}

\bibitem [\protect \citeauthoryear {%
Ward%
}{%
Ward%
}{%
{\protect \APACyear {2013}}%
}]{%
806}
\APACinsertmetastar {%
806}%
\begin{APACrefauthors}%
Ward, J.%
\end{APACrefauthors}%
\unskip\
\newblock
\APACrefYearMonthDay{2013}{}{}.
\newblock
{\BBOQ}\APACrefatitle {Synesthesia} {Synesthesia}.{\BBCQ}
\newblock
\APACjournalVolNumPages{Annual review of psychology}{64}{1}{49--75}
\newblock

\newblock

\PrintBackRefs{\CurrentBib}

\bibitem [\protect \citeauthoryear {%
Webb%
}{%
Webb%
}{%
{\protect \APACyear {1993}}%
}]{%
839}
\APACinsertmetastar {%
839}%
\begin{APACrefauthors}%
Webb, B.%
\end{APACrefauthors}%
\unskip\
\newblock
\APACrefYear{1993}.
\newblock
\APACrefbtitle {Modeling Biological Behaviour or" Dumb Animals and Stupid Robots"} {Modeling biological behaviour or" dumb animals and stupid robots"}.
\newblock
\APACaddressPublisher{}{University of Edinburgh, Department of Artificial Intelligence}.
\PrintBackRefs{\CurrentBib}

\bibitem [\protect \citeauthoryear {%
Welling%
}{%
Welling%
}{%
{\protect \APACyear {2007}}%
}]{%
837}
\APACinsertmetastar {%
837}%
\begin{APACrefauthors}%
Welling, H.%
\end{APACrefauthors}%
\unskip\
\newblock
\APACrefYearMonthDay{2007}{}{}.
\newblock
{\BBOQ}\APACrefatitle {Four mental operations in creative cognition: The importance of abstraction} {Four mental operations in creative cognition: The importance of abstraction}.{\BBCQ}
\newblock
\APACjournalVolNumPages{Creativity research journal}{19}{2-3}{163--177}
\newblock

\newblock

\PrintBackRefs{\CurrentBib}

\bibitem [\protect \citeauthoryear {%
Wheeler%
}{%
Wheeler%
}{%
{\protect \APACyear {2008}}%
}]{%
746}
\APACinsertmetastar {%
746}%
\begin{APACrefauthors}%
Wheeler, M.%
\end{APACrefauthors}%
\unskip\
\newblock
\APACrefYearMonthDay{2008}{}{}.
\newblock
{\BBOQ}\APACrefatitle {Cognition in context: phenomenology, situated robotics and the frame problem} {Cognition in context: phenomenology, situated robotics and the frame problem}.{\BBCQ}
\newblock
\APACjournalVolNumPages{International journal of philosophical studies}{16}{3}{323--349}
\newblock

\newblock

\PrintBackRefs{\CurrentBib}

\bibitem [\protect \citeauthoryear {%
Wolpert%
}{%
Wolpert%
}{%
{\protect \APACyear {2013}}%
}]{%
802}
\APACinsertmetastar {%
802}%
\begin{APACrefauthors}%
Wolpert, D.H.%
\end{APACrefauthors}%
\unskip\
\newblock
\APACrefYearMonthDay{2013}{}{}.
\newblock
{\BBOQ}\APACrefatitle {Ubiquity symposium: Evolutionary computation and the processes of life: What the no free lunch theorems really mean: How to improve search algorithms} {Ubiquity symposium: Evolutionary computation and the processes of life: What the no free lunch theorems really mean: How to improve search algorithms}.{\BBCQ}
\newblock
\APACjournalVolNumPages{Ubiquity}{2013}{December}{1--15}
\newblock

\newblock

\PrintBackRefs{\CurrentBib}

\bibitem [\protect \citeauthoryear {%
Wolpert%
\ \BBA {} Macready%
}{%
Wolpert%
\ \BBA {} Macready%
}{%
{\protect \APACyear {1997}}%
}]{%
801}
\APACinsertmetastar {%
801}%
\begin{APACrefauthors}%
Wolpert, D.H.%
\BCBT {}\ \BBA {} Macready, W.G.%
\end{APACrefauthors}%
\unskip\
\newblock
\APACrefYearMonthDay{1997}{}{}.
\newblock
{\BBOQ}\APACrefatitle {No free lunch theorems for optimization} {No free lunch theorems for optimization}.{\BBCQ}
\newblock
\APACjournalVolNumPages{IEEE transactions on evolutionary computation}{1}{1}{67--82}
\newblock

\newblock

\PrintBackRefs{\CurrentBib}

\bibitem [\protect \citeauthoryear {%
Wu%
\ \protect \BOthers {.}}{%
Wu%
\ \protect \BOthers {.}}{%
{\protect \APACyear {2023}}%
}]{%
786}
\APACinsertmetastar {%
786}%
\begin{APACrefauthors}%
Wu, Z.%
, Qiu, L.%
, Ross, A.%
, Aky{\"u}rek, E.%
, Chen, B.%
, Wang, B.%
\BDBL {}Kim, Y.%
\end{APACrefauthors}%
\unskip\
\newblock
\APACrefYearMonthDay{2023}{}{}.
\newblock
{\BBOQ}\APACrefatitle {Reasoning or reciting? exploring the capabilities and limitations of language models through counterfactual tasks} {Reasoning or reciting? exploring the capabilities and limitations of language models through counterfactual tasks}.{\BBCQ}
\newblock
\APACjournalVolNumPages{arXiv preprint arXiv:2307.02477}{}{}{}
\newblock

\newblock

\PrintBackRefs{\CurrentBib}

\bibitem [\protect \citeauthoryear {%
Yong%
}{%
Yong%
}{%
{\protect \APACyear {2022}}%
}]{%
1006}
\APACinsertmetastar {%
1006}%
\begin{APACrefauthors}%
Yong, E.%
\end{APACrefauthors}%
\unskip\
\newblock
\APACrefYear{2022}.
\newblock
\APACrefbtitle {An immense world: How animal senses reveal the hidden realms around us} {An immense world: How animal senses reveal the hidden realms around us}.
\newblock
\APACaddressPublisher{}{Knopf Canada}.
\PrintBackRefs{\CurrentBib}

\end{thebibliography}

\end{document}